\documentclass[journal]{IEEEtran}
\usepackage{cite}
\usepackage{amsmath,amssymb,amsfonts}

\usepackage{graphicx}
\usepackage{textcomp}
\usepackage{xcolor}
\usepackage{url}
\usepackage{booktabs} 
\usepackage{multirow}
\usepackage{multicol} 
\usepackage{subfig}
\usepackage{array}

\ifCLASSINFOpdf
\else
\fi

\begin{document}
%
\title{FAMED-Net: A Fast and Accurate Multi-scale End-to-end Dehazing Network}
%
%
%

\author{Jing Zhang and Dacheng Tao,~\IEEEmembership{Fellow,~IEEE}

\thanks{This work was supported by the National Natural Science Foundation of China (61806062 and 61873077) and Australian Research Council Projects (FL-170100117, DP-180103424, IH-180100002). Corresponding author: D. Tao (dacheng.tao@sydney.edu.au).}

\thanks{J. Zhang and D. Tao are with the UBTECH Sydney Artificial Intelligence Centre and the School of Computer Science, in the Faculty of Engineering, at the University of Sydney, 6 Cleveland St, Darlington, NSW 2008, Australia.
}
}

%
%

\markboth{Journal of \LaTeX\ Class Files,~Vol.~14, No.~8, September~2018}%
{Shell \MakeLowercase{\textit{et al.}}: Bare Demo of IEEEtran.cls for IEEE Journals}
%



\maketitle

\begin{abstract}
Single image dehazing is a critical image pre-processing step for subsequent high-level computer vision tasks. However, it remains challenging due to its ill-posed nature. Existing dehazing models tend to suffer from model overcomplexity and computational inefficiency or have limited representation capacity. To tackle these challenges, here we propose a fast and accurate multi-scale end-to-end dehazing network called FAMED-Net, which comprises encoders at three scales and a fusion module to efficiently and directly learn the haze-free image. Each encoder consists of cascaded and densely connected point-wise convolutional layers and pooling layers. Since no larger convolutional kernels are used and features are reused layer-by-layer, FAMED-Net is lightweight and computationally efficient. Thorough empirical studies on public synthetic datasets (including RESIDE)  and real-world hazy images demonstrate the superiority of FAMED-Net over other representative state-of-the-art models with respect to model complexity, computational efficiency, restoration accuracy, and cross-set generalization. The code will be made publicly available.
\end{abstract}

\begin{IEEEkeywords}
Dehazing, Image Restoration, Point-wise convolution, Deep Neural Network.
\end{IEEEkeywords}

%
\IEEEpeerreviewmaketitle

\section{Introduction}
\label{sec:intro}

Images captured in hazy conditions often suffer from absorption and scattering effects caused by floating atmospheric particles such as dust, mist, and fumes, which can result in low contrast, blurry, and noisy images. This degraded image quality potentially challenges many subsequent high-level computer vision tasks, $e.g.$, object detection \cite{li2017end, liu2018improved, li2018benchmarking} and segmentation \cite{tu2005image,tarel2010improved, sakaridis2018semantic}. Therefore, removing haze and improving image quality benefits these applications, making image dehazing a subject of intense research and practical focus.

To be specific, image haze removal or dehazing refers to a technique that restores a haze-free image from a single or several observed hazy images. Many dehazing approaches have been proposed, which can be categorized into those that: 1) use auxiliary information such as scene depth \cite{tan2000enhancement} and polarization \cite{schechner2001instant}; 2) use a sequence of captured images \cite{nayar1999vision}; 3) use a single hazy image \cite{liu2018single, wang2018aipnet, li2018single, fattal2008single, he2011single, tang2014investigating, zhu2015fast, berman2016non, cai2016dehazenet, ren2016single, li2017all, Zhang2018Fully, Ren_2018_CVPR, zhang2018densely, Li_2018_CVPR, bui2018single}, as the model input when dehazing. Of these, single image dehazing without the need for additional information is of most practical benefit. However, as a typical ill-posed problem, single image dehazing remains challenging and requires refinement.

The presence of haze leads to the combination of an attenuation term corresponding to the absorbing effect and a scattering term corresponding to the scattering effect that occur during imaging. Both terms are related to an intermediate variable, that is, transmission, which depends on scene depth. One feasible haze removal solution is to estimate the transmission and then recover the clear image by reversing the attenuation and scattering. Many single image dehazing methods have been proposed \cite{fattal2008single, CVPR2009_He, he2011single, tang2014investigating, zhu2015fast, berman2016non, cai2016dehazenet, Zhang2018Fully}, which use either hand-crafted features ($e.g.$, different image priors) or learning-based features to estimate the haze transmission.

For example, He \textit{et al.} \cite{he2011single} proposed a simple and effective dark channel prior for single image dehazing, which assumes that the minimum of all the spectral channels in clear images (the ``dark channel'') is close to zero. The method effectively estimates the haze transmission. However, the dark channel prior may not work for some particular scenes such as for white objects, which are similar to atmospheric light, because it underestimates the transmission and leads to over-dehazed artifacts. Zhu \textit{et al.} \cite{zhu2015fast}  proposed a color attenuation prior that assumes a positive correlation between the scene depth and the haze concentration, which is represented by the subtraction of scene brightness from saturation. Then, the scene depth and haze transmission are easily estimated by a regressed linear model based on the above prior. Recently, Berman \textit{et al.} \cite{berman2016non} proposed a non-local prior based on the assumption that colors in a clear image can be approximated by some distinct colors clustering tightly in RGB space. Being affected by haze, each cluster becomes a line in RGB space (haze-line) due to the varying transmission coefficients of the clustered pixels. Consequently, the transmission and clear image are estimated according to these haze lines. Though prior-based methods are usually simple and effective for many scenes, they share the common limitation of describing specific statistics, which may not work for some images.

Learning-based methods adopt a data-driven approach to learn a linear/non-linear mapping between features and transmission and so overcomes the limitations of specific priors. For example, Tang \textit{et al.} \cite{tang2014investigating} proposed learning a regression model based on random forests from haze-relevant features including the dark channel, local max contrast, hue disparity, and local max saturation. They trained the model using a synthetic dataset and tested it on both synthetic and real-world hazy images, which then became common practice in subsequent learning-based methods. The learning-based idea for dehazing has subsequently been extended in three ways: 1) more powerful learning models; 2) more effective synthetic methods and larger datasets; 3) end-to-end modeling/training.

Deep neural networks have now been successfully applied to many computer vision tasks including object recognition, detection, and semantic segmentation. By leveraging their powerful representation capacity and end-to-end learning, many deep convolutional neural network (CNN)-based approaches were proposed for image dehazing \cite{cai2016dehazenet, ren2016single, li2017all, Zhang2018Fully, Ren_2018_CVPR, zhang2018densely, Li_2018_CVPR}. For example, Cai \textit{et al.} \cite{cai2016dehazenet} proposed an end-to-end trainable deep CNN model called DehazeNet to directly learn the transmission from hazy images, which is superior to contemporary prior-based methods and random forest models \cite{tang2014investigating}. Ren \textit{et al.} \cite{ren2016single} proposed a multi-scale CNN (MSCNN) to learn the transmission map in a fully convolutional manner and explore a multi-scale architecture for coarse-to-fine regression.

Despite the effectiveness of CNN-based approaches, a separate step is still needed to estimate the atmospheric light. Recently, Zhang \textit{et al.} \cite{zhang2018densely} proposed an end-to-end densely connected pyramid dehazing network (DCPDN) to jointly learn the transmission map, atmospheric light, and dehazing. They adopted an encoder-decoder architecture with a multi-level pyramid pooling module to learn multi-scale features. They also utilized an adversarial loss based on a generative adversarial network \cite{goodfellow2014generative} to supervise the dehazing network. Rather than estimating the intermediate transmission, Li et al .\cite{li2017all} proposed an end-to-end CNN model called the all-in-one dehazing network (AOD-Net) to learn the clear image from a hazy one. They integrated the transmission and atmospheric light into a single variable by reformulating the hazy imaging model. Ren \textit{et al.} \cite{Ren_2018_CVPR} proposed a gated fusion network (GFN) by adopting an encoder-decoder architecture, while Li \textit{et al.}\cite{Li_2018_CVPR} also designed an encoder-decoder architecture but based on a conditional generative adversarial network (cGAN) to learn the dehazed image end-to-end. Though cGAN and DCPDN have achieved good dehazing results, they contain dozens of convolutional layers and are about 200 MB in size, making them awkward and unlikely to be applicable in the resource-constrained context of a computer vision system.

In this paper, we aim to develop a fast and accurate deep CNN model for single image dehazing. We use a fully convolutional and end-to-end training/testing approach to efficiently process hazy images of arbitrary size. To this end, we propose a fast and accurate multi-scale dehazing network called FAMED-Net, which comprises encoders at three scales and a fusion module to directly learn the haze-free image. Each encoder consists of cascaded point-wise convolutional layers and pooling layers via a densely connected mechanism. Since no larger convolutional kernels are used and features are reused layer-by-layer, FAMED-Net is lightweight and computationally efficient. Thorough empirical studies on public synthetic datasets and real-world hazy images demonstrate the superiority of FAMED-Net over representative state-of-the-art models with respect to model complexity, computational efficiency, restoration accuracy, and cross-set generalization. The code will be made publicly available at \url{https://github.com/chaimi2013/FAMED-Net}.

The main contributions of this paper can be summarized as follows:

$\bullet$ We devise a novel multi-scale end-to-end dehazing network called FAMED-Net, which implicitly learns efficient statistical image priors for fast and accurate haze removal from a single image.

$\bullet$ FAMED-Net leverages fully point-wise convolutions as the basic unit to construct the encoder-decoder architecture, which has a small model size and is computationally efficient.

$\bullet$ FAMED-Net outperforms state-of-the-art models on both synthetic benchmarks and real-world hazy images.


\section{Related Work}
\label{sec:relatedWork}
\subsection{Atmospheric Scattering Model}

Images captured in hazy condition can be mathematically formulated as \cite{mccartney1976optics, narasimhan2002vision, he2011single}:
\begin{equation}
{I^\lambda }\left( x \right) = {J^\lambda }\left( x \right)t\left( x \right) + {A^\lambda }\left( {1 - t\left( x \right)} \right),
\label{eq:HazyModel}
\end{equation}
where $I$ is the observed hazy image, $J$ is the scene radiance, $A$ is the atmospheric light assumed to be a global constant, $t$ is the haze transmission, $x$ denotes pixel location, and $\lambda$ denotes the spectral channel, $i.e.$, $\lambda  \in \left\{ {r,g,b} \right\}$. The first term, called the attenuation term, represents the haze absorbing effect on scene radiance, while the second term, called the scattering term, represents the haze scattering effect on ambient light. $t$ describes the fraction of scene radiance reaching the camera sensor, so is the ``transmission'', which depends on scene depth. Under the homogeneous haze assumption, the transmission can be expressed as:
\begin{equation}
t\left( x \right) = {e^{ - \beta d\left( x \right)}},
\label{eq:t}
\end{equation}
where $\beta$ denotes the medium attenuation coefficient and $d$ is the scene depth.

Recently, Li \textit{et al.} \cite{li2017all} reformulated the imaging model in  Eq.~\eqref{eq:HazyModel} by integrating the transmission and atmospheric light into a single variable $K$:
\begin{equation}
{K^\lambda }\left( x \right) \buildrel \Delta \over = \frac{{\frac{1}{{t\left( x \right)}}\left( {{I^\lambda }\left( x \right) - {A^\lambda }} \right) + \left( {{A^\lambda } - 1} \right)}}{{{I^\lambda }\left( x \right) - 1}},
\label{eq:K}
\end{equation}
\begin{equation}
{J^\lambda }\left( x \right) = {K^\lambda }\left( x \right){I^\lambda }\left( x \right) - {K^\lambda }\left( x \right) + 1.
\label{eq:KModel}
\end{equation}
They designed an end-to-end network (AOD-Net) which learns a direct mapping from a raw hazy image to scene radiance.

\subsection{Prior-based and Learning-based Image Dehazing Methods}
\label{subsec:related_work_dehazing_methods}
\begin{figure}
\centering
\includegraphics[width=1\linewidth]{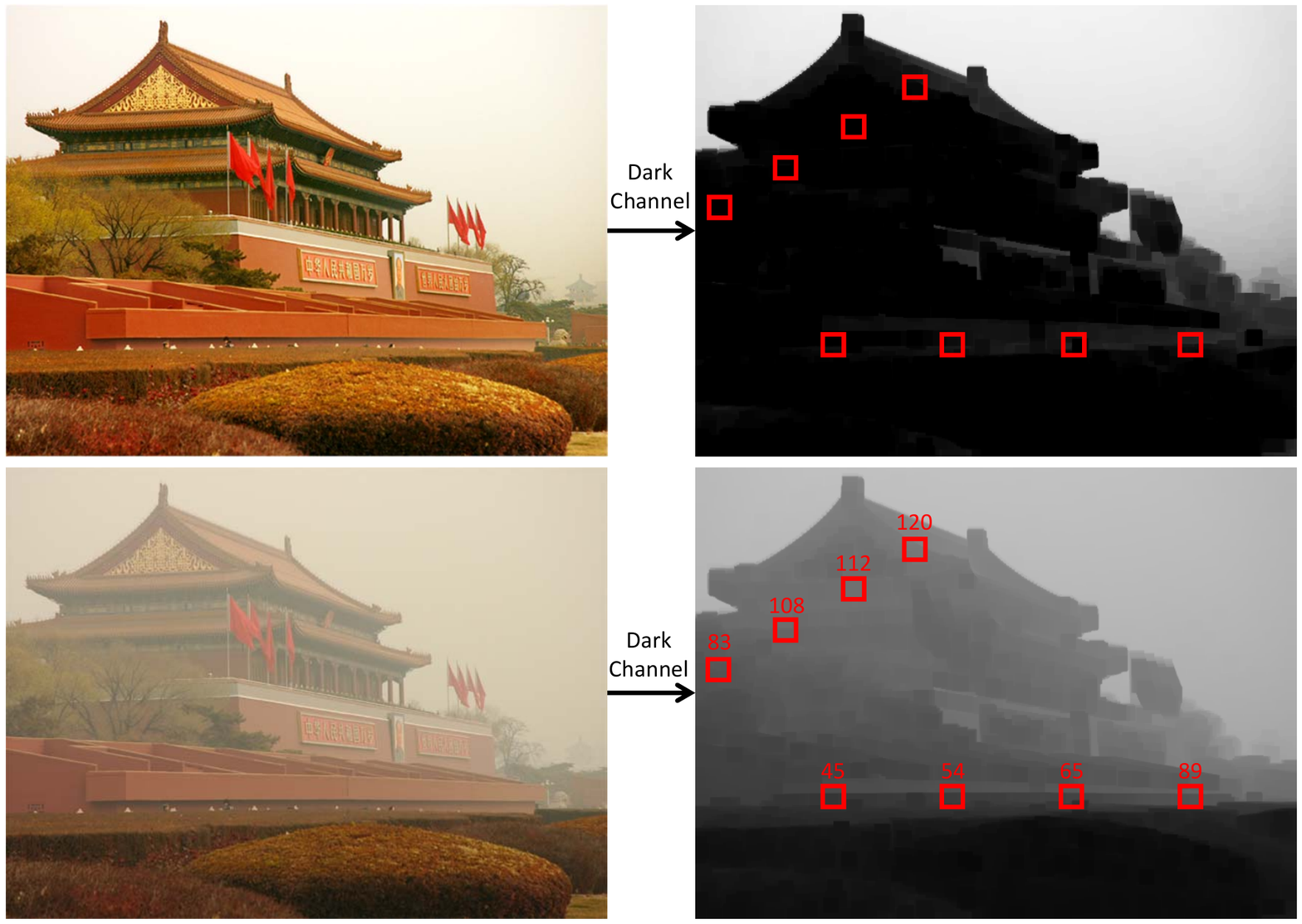}
\caption{An illustration of dark channels calculated on both clear (the top row) and hazy images (the bottom row). }
\label{fig:dcp}
\end{figure}

As can be seen from the atmospheric scattering model in Eq.~\eqref{eq:HazyModel}, given an observed hazy image $I$, recovering the scene radiance is ill-posed. Different image priors have been proposed to constrain the haze-free image and make the estimate tractable, including the dark channel prior \cite{he2011single}, color attenuation prior \cite{zhu2015fast}, and non-local prior \cite{berman2016non}, etc. As defined in \cite{he2011single}, each pixel value of the dark channel refers to the minimum pixel value on each patch centered at every pixel position. Figure~\ref{fig:dcp} shows an example of the dark channels on both clear and hazy images. As can be seen, the dark channel of a clear image is almost dark everywhere except for the bright sky region, while the dark channel of a hazy image reveals the haze veil due to the haze scattering effect (corresponds to the second term in Eq.~\eqref{eq:HazyModel}). Based on the dark channel prior, the transmission can be efficiently estimated from the dark channel map. It is noteworthy that the pixel value of the dark channel reveals the hazy density (which is related to scene depth) even though it is calculated locally in a sliding window manner (See the regions and corresponding values indicated by the red boxes). It can be explained as follows: 1) the haze effects of both attenuation and scattering which are directly related to scene depth, can be described as a pixel-to-pixel ($i.e.$, locally) mapping from clear pixel to hazy pixel by the atmospheric scattering model. 2) the dark channel prior reveals the intrinsic locally statistical property of clear images. Similar to \cite{he2011single}, our approach also solves the dehazing problem in a local manner which implicitly learns a statistical image prior as will be demonstrated in Section~\ref{subsubsec:exp_comparisonsStatistical}.

To overcome the limitations of prior-based methods, many deep CNN-based data-driven dehazing models have been proposed since Cai \textit{et al.} \cite{cai2016dehazenet} proposed DehazeNet, including MSCNN \cite{ren2016single}, AOD-Net \cite{li2017all}, FPCNet \cite{Zhang2018Fully}, DCPDN \cite{zhang2018densely}, GFN \cite{Ren_2018_CVPR}, cGAN \cite{Li_2018_CVPR} and proximal DehazeNet \cite{Yang_2018_ECCV}. These can be categorized into those that: 1) estimate $t$ using CNN \cite{cai2016dehazenet, ren2016single, Zhang2018Fully, zhang2018densely, Yang_2018_ECCV}; 2) directly learn the scene radiance end-to-end \cite{li2017all, zhang2018densely, Ren_2018_CVPR, Li_2018_CVPR, Yang_2018_ECCV}. Our proposed method falls into the latter category and is partly inspired by AOD-Net \cite{li2017all} and FPCNet \cite{Zhang2018Fully}. In contrast to AOD-Net, we propose a fully point-wise CNN to regress $K$ and produce a stronger representation capacity. In contrast to FPCNet, we propose: 1) an end-to-end model to regress the scene radiance directly; 2) a multi-scale architecture to handle the scale variance, which achieves much better results than FPCNet while maintaining low model complexity and high computational efficiency; and 3) a new training/testing strategy that negates the need for a pre-processing shuffling step. Compared to MSCNN, in which coarse-scale predictions are used as part of the input for the finer scale, the proposed method adopts a Gaussian pyramid architecture and follows a late fusion strategy. It produces better dehazing results than MSCNN and runs faster. Compared to the recently proposed DCPDN and cGAN, our model is much more compact, $i.e.$, less than 90 kb vs. about 200 Mb, while having a high restoration accuracy and computational efficiency.

\begin{figure*}
\centering
\includegraphics[width=1\linewidth]{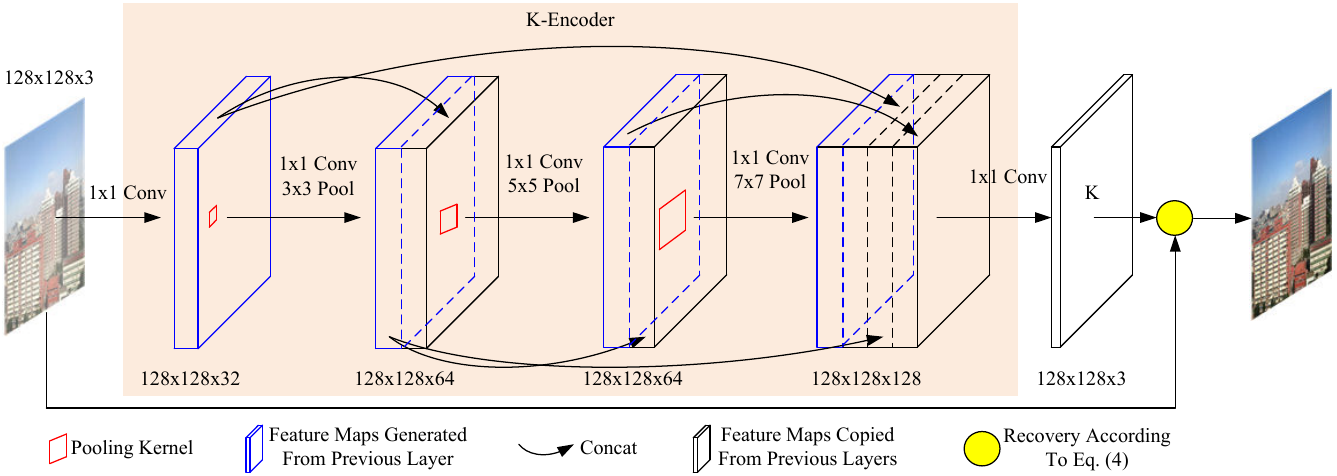}
\caption{Schematic of the end-to-end fully point-wise CNN for single image dehazing, $i.e.$, FAMED-Net-SS. K-encoder comprises cascaded point-wise convolutional layers and pooling layers via a dense connected mechanism for learning $K$ in Eq.~\eqref{eq:K}.}
\label{fig:network}
\end{figure*}

\subsection{Multi-scale pyramid architecture}
The pyramid structure is a basic idea used for both multi-resolution image representation and multi-scale feature representation in the computer vision area, for example, Gaussian pyramid, Laplacian pyramid, wavelet \cite{daubechies1990wavelet}, and SIFT \cite{lowe2004distinctive}. Leveraging this classical idea, CNN produces a feature pyramid through stacked convolutional layers and spatial pooling layers. Recently, different multi-scale image or feature pyramid architectures have been devised for both low- and high-level computer vision applications, including deep Laplacian pyramid networks for image super-resolution \cite{lai2017deep}, DeepExposure using Laplacian pyramid decomposition \cite{yu2018deepexposure}, deep generative image models \cite{denton2015deep}, Laplacian pyramid reconstructive adversarial network \cite{xu2018lapran}, Deeplab using an image pyramid for semantic segmentation \cite{chen2018deeplab}, and feature pyramid networks for object detection \cite{Lin_2017_CVPR}. Our approach also adopts the Gaussian/Laplacian pyramid architectures for multi-scale fusion (See Figure~\ref{fig:network_multiscale_fusion} and Figure~\ref{fig:network_multiscale_residual}). In contrast to those above methods, the proposed FAMED-Net is specifically devised for single image dehazing. Moreover, it leverages fully point-wise convolutions instead of convolutions with large kernels for constructing a lightweight and computationally efficient network.

\subsection{Deep Supervision}
Adding auxiliary supervision on intermediate layers within a deep neural network also known as \emph{deep supervision} is originally proposed by Xie and Tu in the seminal work \cite{xie2015holistically,lee2015deeply}. This technique facilitates multi-scale and multi-level feature learning by allowing error information backpropagation from multiple paths and alleviating the problem of vanishing gradients in deep neural networks. Deep supervision has been widely adopted in the following work in different areas such as Deeplab for semantic segmentation \cite{chen2018deeplab}, MSCNN for image dehazing \cite{ren2016single}, LapSRN for image super-resolution \cite{lai2017deep}, etc. We also add supervision on the dehazed image at each scale by leveraging the deep supervision idea.

\section{FAMED-Net for Single Image Dehazing}
\label{sec:proposedMethod}


\subsection{A Probabilistic View to Solving the Ill-posed Dehazing Problem}
\label{subsec:ProbabilisticView}
Eq.~\eqref{eq:HazyModel} and Eq.~\eqref{eq:KModel} can be re-written as:
\begin{equation}
\left( {{I^\lambda }\left( x \right) - {A^\lambda }} \right) = \left( {{J^\lambda }\left( x \right) - {A^\lambda }} \right)t\left( x \right),
\label{eq:hazyModelReform}
\end{equation}
\begin{equation}
\left( {{I^\lambda }\left( x \right) - 1} \right) = \left( {{J^\lambda }\left( x \right) - 1} \right)\frac{1}{{{K^\lambda }\left( x \right)}}.
\label{eq:kModelReform}
\end{equation}
Applying a logarithmic operation to both sides of the above equation produces the following general form:
\begin{equation}
y = x + z,
\label{eq:illPosedModel}
\end{equation}
where $y$ is the observed degraded image, $x$ is the ground truth haze-free image, and $z$ is the intermediate variable related to the degrading process. $x$ and $z$ can be estimated using maximum a posteriori estimation (MAP), $i.e.$,
\begin{align}\nonumber
 \left( {{x^*},{z^*}} \right) &= \mathop {\arg \max }\limits_{\left( {x,z} \right)} p\left( {x,z\left| y \right.} \right) \\ \nonumber
  &= \mathop {\arg \max }\limits_{\left( {x,z} \right)} \frac{{p\left( {y\left| {x,z} \right.} \right)p\left( {x,z} \right)}}{{\int\limits_X {\int\limits_Z {p\left( {y\left| {x,z} \right.} \right)p\left( {x,z} \right)dxdz} } }} \\
  &= \mathop {\arg \max }\limits_{\left( {x,z} \right)} p\left( {y\left| {x,z} \right.} \right)p\left( {z\left| x \right.} \right)p\left( x \right).
 \label{eq:MAP}
\end{align}
$p\left( {y\left| {x,z} \right.} \right)$ is the data likelihood, which corresponds to the data fidelity term measuring the reconstruction error. When using the L2 loss to supervise network training, it indeed assumes a normal distribution about the reconstruction error (see Section~\ref{subsec:network} and the yellow circle in Figure~\ref{fig:network}). The L1 loss can also be used to enforce a sparse constraint. $p\left( {z\left| x \right.} \right)$ is the conditional distribution of $z$ conditioned on the clear haze-free image. For example, DCP \cite{he2011single} assumes that $p\left( {DarkChannel\left| x \right.} \right)$ ($i.e.$, $p\left( {1 - t\left| x \right.} \right)$) concentrates on zeros. As with DehazeNet \cite{cai2016dehazenet} and AOD-Net \cite{li2017all}, the networks can implicitly learn $p\left( {t\left| x \right.} \right)$ and $p\left( {K\left| x \right.} \right)$, as we will show in Section~\ref{subsubsec:exp_comparisonsStatistical}. $p\left( x \right)$ is the prior distribution of $x$, usually assumed to be long-tailed due to the spatial continuity in natural images (locally smooth regions and sparse abrupt edges) \cite{Aapo2009Natural}. Markov Random Fields or simple filters like guided filter are used to model the spatial continuity \cite{he2013guided}.

Based on the above analysis, the key is to construct a model that can effectively learn statistical regularities. As shown in \cite{Zhang2018Fully}, statistical regularities in natural images can be efficiently learned by point-wise convolutions, which are compact and resists over-fitting. Partly inspired by \cite{Zhang2018Fully}, we devise a novel end-to-end fully point-wise CNN for single image dehazing.

\begin{figure*}
\centering
\subfloat[]{\label{fig:network_multiscale_fusion}
\includegraphics[width=0.49\linewidth]{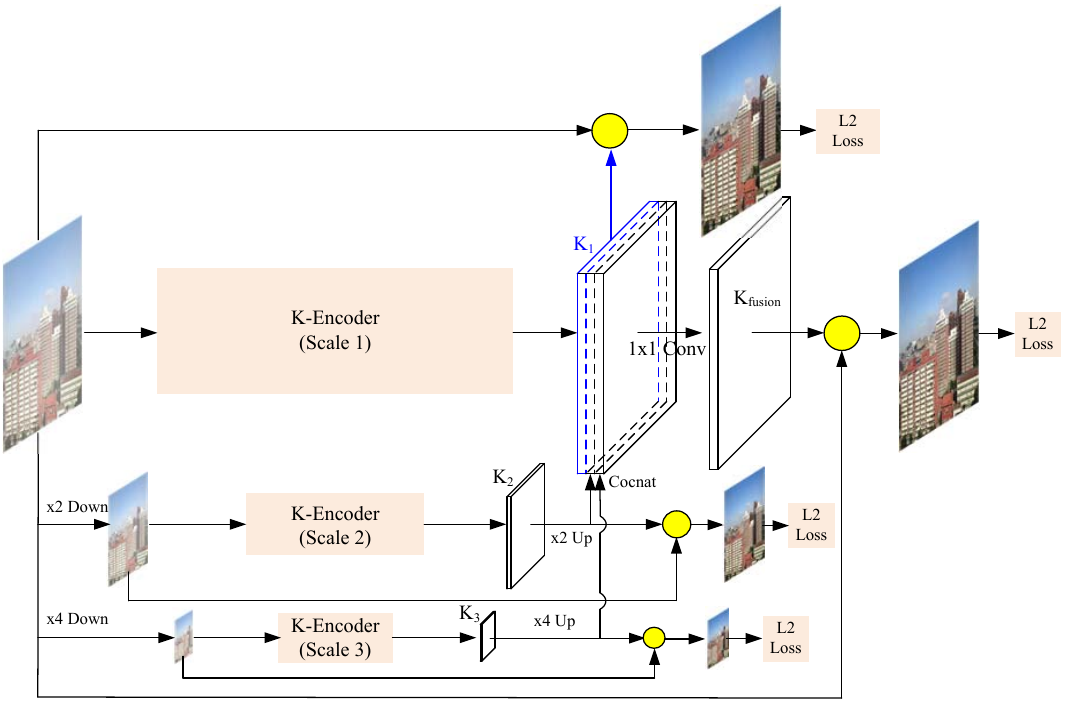}}
\hspace{0.01\linewidth}
\subfloat[]{\label{fig:network_multiscale_residual}
\includegraphics[width=0.49\linewidth]{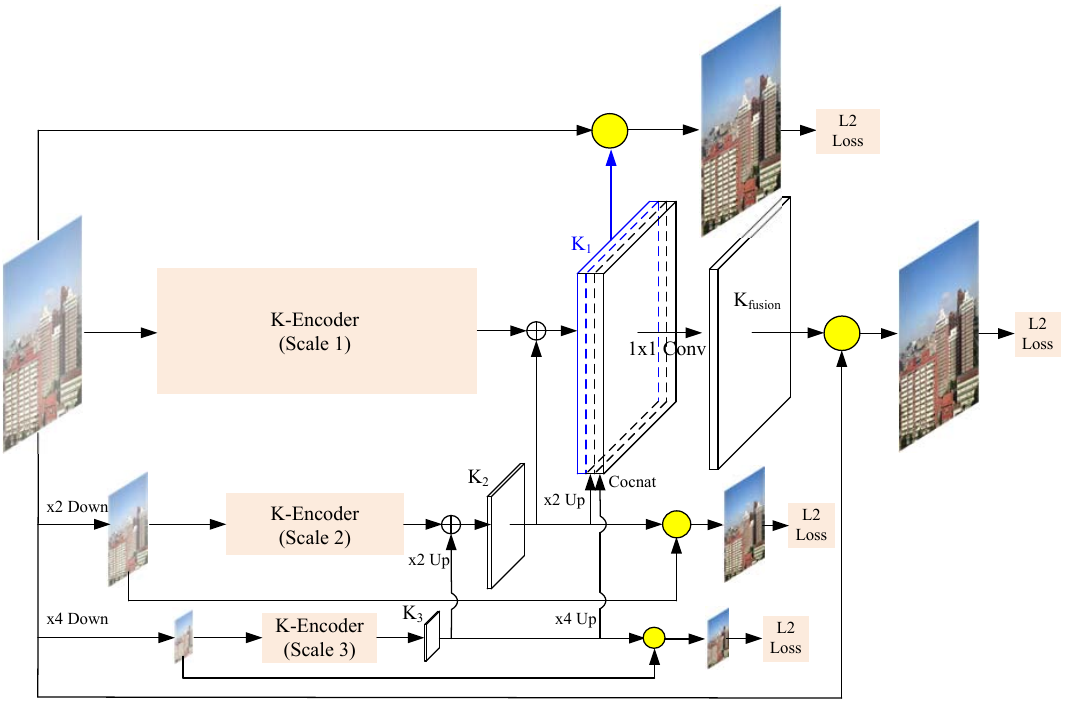}}

\caption{Schematic of the multi-scale FAMED-Net architecture. (a) The Gaussian pyramid architecture with a late fusion module, $i.e.$, FAMED-Net-GP. (b) The Laplacian pyramid architecture with a late fusion module, $i.e.$, FAMED-Net-LP.}
\label{fig:network_multiscale}
\end{figure*}

\subsection{The Single-scale FAMED-Net: FAMED-Net-SS}
\label{subsec:network}
As shown in Figure~\ref{fig:network}, the network is designed to learn the reformulated variable $K$ in Eq.~\eqref{eq:K} and recover the scene radiance according to Eq.~\eqref{eq:KModel} (see \cite{li2017all}). There are five point-wise convolutional layers, in which the first four form the K-encoder and the last forms the decoder. Features corresponding to different receptive fields are reused via dense connections (see black arcs and cubes in Figure~\ref{fig:network}). Mathematically, this can be formulated as:
\begin{equation}
{f^{l + 1}} = {\varphi ^{l + 1}}\left( {concat\left( {{f^k}\left| {k \in {\Lambda ^{l + 1}}} \right.} \right)} \right),l \in \left[ {0,4} \right],
\label{eq:featInFamedNetSS}
\end{equation}
where ${{f^k}}$ represents the learned features from the $k^{th}$ block. We denote the input as the $0^{th}$ block, the hazy image of size $H \times W \times 3$ as ${{f^0}}$, and the decoded features in the $5^{th}$ block as $K$, $i.e.$, $K \buildrel \Delta \over = {f^5}$. ${{\Lambda ^{l + 1}}}$ denotes the index set, which indexes the feature maps used by the $(l+1)^{th}$ block via dense connections ($concat$), $i.e.$, ${\Lambda ^1} = \{ 0\}$, ${\Lambda ^2} = \{ 1\}$, ${\Lambda ^3} = \{ 1,2\}$, ${\Lambda ^4} = \{ 2,3\}$, ${\Lambda ^5} = \{ 1,2,3,4\}$ in the proposed network. ${\varphi ^{l + 1}}$ denotes the mapping function in the $(l+1)^{th}$ block learned by a combination of a convolutional layer, a batch normalization layer, a ReLU layer and a pooling layer.

We leverage pooling layers of different kernel sizes ($r^l \times r^l$) after each convolutional layer to aggregate multi-level statistics (features) within the receptive fields, $i.e.$, $r^l = 2l - 1,l \in \left[ {1,4} \right]$. It is noteworthy that by using a combination of point-wise convolutional layers and a $r^l \times r^l$ pooling layer, the output node has a receptive field of $r^l \times r^l$, which is equivalent to the one using a $r^l \times r^l$ convolutional layer alone. In this way, we retain the representation capacity of the neural network for statistical modeling but using fewer parameters, leading to a more compact architecture. Further, no pooling layer and batch normalization layer are used in the final $5^{th}$ block. Since pooling with a $1\times1$ kernel is trivial, it is omitted. Strides in both the convolutional and pooling layers are set to 1 to retain the feature map size. The output feature channels in the K-encoder are kept at 32, $i.e.$, ${f^l} \in {R^{H \times W \times 32}},l \in \left[ {1,4} \right]$ (see blue cubes in Figure~\ref{fig:network}). Then, the decoded $K$ map is used to recover the scene radiance according to Eq.~\eqref{eq:KModel} (see the yellow circle in Figure~\ref{fig:network}). This structure is denoted FAMED-Net-SS, where ``SS'' stands for single scale.

We use the L2 loss to supervise the network during training:
\begin{equation}
{w^*} = \mathop {\arg \min }\limits_w  {\left\| {J - J\left( {I;w } \right)} \right\|^2} + \lambda {\left\| w \right\|^2},
\label{eq:L2_loss}
\end{equation}
where ${J\left( {I;w } \right)}$ is the estimated scene radiance, $w$ represents learnable parameters of the network, and $\lambda$ is the weight decay factor in the regularization term.

\subsection{The Multi-scale Variants of FAMED-Net: FAMED-Net-GP and FAMED-Net-LP}
\label{subsec:MSNetwork}
Objects at distinct distances are of different size in the captured images, leading to variably sized homogenous regions in the transmission map or $K$ map. To handle the multi-scale characteristics, we extend the proposed network to multi-scale by adopting a Gaussian pyramid architecture as shown in Figure~\ref{fig:network_multiscale}(a). We down-sample the input hazy image to another two scales, $i.e.$, 1/2 scale and 1/4 scale, respectively. Then, we construct a K-encoder for each scale without sharing weights. Further, the estimated $K$ maps from the coarse scales are interpolated to the original scale and concatenated as:
\begin{equation}
{K_{concat}} \buildrel \Delta \over = \left[ {{K_1};{K_2}{ \uparrow _{ \times 2}};{K_3}{ \uparrow _{ \times 4}}} \right],
\label{eq:kConcat}
\end{equation}
where ${K_s}{ \uparrow _{ \times m}},s \in \left[ {2,3} \right],m = 2\left( {s - 1} \right)$ denote the interpolated $K$ maps. Bilinear interpolation is used for both down-sampling and up-sampling. Then, we introduce a fusion module to fuse the multi-scale estimates into a more reliable one, which is again implemented by a $1\times1$ convolutional layer and a ReLU layer ${\varphi ^6}$ as:
\begin{equation}
{K_{fusion}} = {\varphi ^6}\left( {{K_{concat}}} \right).
\label{eq:kFusion}
\end{equation}
Finally, $K_{fusion}$ is used to recover the scene radiance according to Eq.~\eqref{eq:KModel}. This structure is denoted FAMED-Net-GP, where ``GP'' stands for Gaussian pyramid.

The L2 loss is used to supervise the network:
\begin{equation}
\begin{array}{c}
 {w^*} = \mathop {\arg \min }\limits_w \sum\limits_{s = 1,2,3} {{\alpha _s}{{\left\| {{J_s} - {J_s}\left( {I;w} \right)} \right\|}^2}}  +  \\
 \qquad \qquad {\alpha _{fusion}}{\left\| {{J_1} - {J_{fusion}}\left( {I;w} \right)} \right\|^2} + \lambda {\left\| w \right\|^2} \\
 \end{array},
\label{eq:MS_L2_loss}
\end{equation}
where $J_s$ and ${{J_s}\left( {I;w} \right)}$ represent the ground truth and the estimated scene radiance at each scale, and ${{J_{fusion}}\left( {I;w} \right)}$ represents the estimated scene radiance from the fusion module. $\alpha_s$ and $\alpha _{fusion}$ are loss weights, which are set to 1.

In addition to the Gaussian pyramid architecture, we also adopt a Laplacian pyramid architecture for comparison. As shown in Figure~\ref{fig:network_multiscale}(b), the estimated $K$ map at the coarse scale is interpolated and added to the K-encoder output at the finer scale. Mathematically, it can be formulated as:
\begin{equation}
{K_s} = {K_{s + 1}}{ \uparrow _{ \times 2}} + \Delta {K_s},s \in \left[ {1,2} \right].
\label{eq:kResidual}
\end{equation}
Therefore, it enforces the K-encoder at the finer $s^{th}$ scale to learn a residual $\Delta {K_s}$. The other parts are kept the same as the Gaussian pyramid one. This structure is denoted FAMED-Net-LP, where ``LP'' stands for Laplacian pyramid. It is noteworthy that the receptive field of FAMED-Net-SS is $13\times13$ which is similar to the local window size in prior-based dehazing methods, $e.g.$, $15\times15$ in DCP \cite{CVPR2009_He} and MRP \cite{zhang2017fast}. As for FAMED-Net-GP and FAMED-Net-LP, their receptive fields become larger, $i.e.$, $52\times52$, which enables the network to learn more effective statistical regularities.

\subsection{Model Complexity Analysis}
\label{subsec:ComplexityAnalysis}

\begin{table}[htbp]
\footnotesize
  \newcommand{\tabincell}[2]{\begin{tabular}{@{}#1@{}}#2\end{tabular}}
  \centering
  \caption{Network architectures of FAMED-Net.}
    \begin{tabular}{p{1.75cm}<{\centering}p{1.15cm}p{1.25cm}<{\centering}p{0.3cm}<{\centering}p{0.4cm}<{\centering}p{0.2cm}<{\centering}p{0.4cm}<{\centering}}
    \toprule
    Network & Type  & Input Size & Num   & Filter & Pad   & Stride \\
    \midrule
    \multirow{12}[0]{*}{FAMED-Net-SS} & Conv1 & 128x128x3 & 32    & 1x1   & 0     & 1 \\
          & Conv2 & 128x128x32 & 32    & 1x1   & 0     & 1 \\
          & Pool2 & 128x128x32 & -     & 3x3   & 1     & 1 \\
          & Concat2 & 128x128x64 & -     & -   & -     & - \\
          & Conv3 & 128x128x64 & 32     & 1x1   & 0     & 1 \\
          & Pool3 & 128x128x32 & -     & 5x5   & 2     & 1 \\
          & Concat3 & 128x128x64 & -     & -   & -     & - \\
          & Conv4 & 128x128x64 & 32     & 1x1   & 0     & 1 \\
          & Pool4 & 128x128x32 & -     & 7x7   & 3     & 1 \\
          & Concat4 & 128x128x128 & -     & -   & -     & - \\
          & Conv5 & 128x128x128 & 3     & 1x1   & 0     & 1 \\
          \cmidrule {2-7}
          & Params & \multicolumn{5}{c}{5,987}               \\
          & Complexity\footnotemark[1] & \multicolumn{5}{c}{9.39x10$^7$}        \\
    \midrule
    FAMED-Net & Params & \multicolumn{5}{c}{17,991}              \\
       (GP/LP)   & Complexity & \multicolumn{5}{c}{1.24x10$^8$}        \\
    \bottomrule
    \end{tabular}%
  \label{tab:FAMED-Net_architecture}%
\end{table}%

\footnotetext[1]{Evaluated with FLOPs, $i.e.$ the number of floating-point multiplication-adds.}

The details of FAMED-Net are shown in Table~\ref{tab:FAMED-Net_architecture}. It can be seen that FAMED-Net is very lightweight and compact thanks to the fully point-wise convolutions. For example, FAMED-Net-SS only contains 5,987 learnable parameters and has 9.39x10$^7$ FLOPs. The number of learnable parameters increases threefold in FAMED-Net-GP, while the FLOPs only increase by about 30\%. FAMED-Net can process hazy images of arbitrary size due to its fully convolutional structure, with the computational cost increases linearly with the image size.

To reduce the required FLOPs for large images, we propose a fixed size testing strategy. First, we resize the hazy image with the longest side to 360 and input it into the network. Then, we resize the estimated $K$ map from the fusion module back to the original size using bilinear interpolation. Further, we use the fast-guided filter \cite{he2015fast} to refine the interpolated $K$ map. The fast-guided filter is $d^2$-times faster than the original $O(N)$-guided filter \cite{he2013guided}, with almost no visible degradation, where $d$ is the down-sampling ratio (refer to \cite{he2015fast} for details). Finally, the scene radiance is recovered according to Eq.\eqref{eq:KModel}. In this way, we can process hazy images of arbitrary size at an almost fixed computational cost. We present our comparisons with state-of-the-art models in Table~\ref{tab:SOA_runningTime} including parameters, model size, and runtime. These comparisons clearly show that FAMED-Net is lightweight and computationally efficient. More details can be found in Section~\ref{subsubsec:exp_comparisonsSOA_runningTime}.

\begin{table}[htbp]
  \centering
  \caption{Comparison of FAMED-Net and state-of-the-art models with respect to parameters, model size, and runtime.}
    \begin{tabular}{p{1.8cm}p{0.85cm}p{0.8cm}p{1.6cm}p{1.7cm}} 
    \toprule
     Model  & Param. & Size &Platform & Time (second) \\
    \midrule
    DCP \cite{CVPR2009_He}   & - & - & Matlab(C) & 1.62 \\
    FVR \cite{tarel2009fast} & - & - & Matlab(C) & 6.79 \\
    BCCR \cite{meng2013efficient} & - & - & Matlab(C) & 2.85 \\
    GRM \cite{chen2016robust} & - & - & Matlab(C) & 83.96\\
    CAP \cite{zhu2015fast} & - & - & Matlab(C) & 0.95\\
    NLD \cite{berman2016non} & - & - & Matlab(C) & 9.89\\
    \midrule
    DehazeNet \cite{cai2016dehazenet} & 8,240 & - & Matlab(C) & 1.3399\\
    MSCNN \cite{ren2016single} & 8,014 & - & Matlab(C) & 2.4840\\
    \multirow{2}[0]{*}{FPCNet \cite{Zhang2018Fully}} & \multirow{2}[0]{*}{288} & \multirow{2}[0]{*}{2.2Kb} & \multirow{2}[0]{*}{MatCaffe(C/G)} & 0.1924/0.0016\\
                & &  &  & 0.2046/0.0178\\
    AOD-Net \cite{li2017all} & 1,833 & 8.9Kb & MatCaffe(C/G) & 0.3025/0.0043\\
    GFN \cite{Ren_2018_CVPR} & 514,415 & 1.99Mb & MatCaffe(C/G) & 9.9763/0.0490\\
    cGAN \cite{Li_2018_CVPR} & 1.23x$10^8$ & 198.8Mb & Torch7(G) & 0.0520\\
    DCPDN\footnotemark[2] \cite{zhang2018densely} & 6.69x$10^7$ & 255.6Mb & Pytorch(G) & 0.0417\\
    \multirow{2}[0]{*}{\textbf{FAMED-Net}} & \multirow{2}[0]{*}{17,991} & \multirow{2}[0]{*}{86.3Kb} & \multirow{2}[0]{*}{MatCaffe(C/G)} & 0.8894/0.0116\\
                  & &  &  & 0.9061/0.0285\\
    \bottomrule
    \end{tabular}%
  \label{tab:SOA_runningTime}%
\end{table}%

\footnotetext[2]{The number was calculated on 512x512 images since DCPDN required a fixed-size input.}

\section{Experiments}
\label{sec:experiments}

To evaluate the performance of FAMED-Net, we compared it with state-of-the-art image prior-based methods including DCP \cite{CVPR2009_He}, FVR \cite{tarel2009fast}, BCCR \cite{meng2013efficient}, GRM \cite{chen2016robust}, CAP \cite{zhu2015fast}, and NLD \cite{berman2016non} and deep CNN-based methods including DehazeNet \cite{cai2016dehazenet}, MSCNN \cite{ren2016single}, AOD-Net \cite{li2017all, liu2018improved}, FPCNet \cite{Zhang2018Fully}, GFN \cite{Ren_2018_CVPR}, and DCPDN \cite{zhang2018densely}. We adopted the recently proposed RESIDE \cite{li2018benchmarking} as the benchmark dataset due to its large scale and diverse data sources and image contents. RESIDE contains 110,500 synthetic hazy indoor images (ITS) and 313,950 synthetic hazy outdoor images (OTS) in the training set. We reported the PSNR and SSIM for each method on the SOTS test set, which includes both indoor and outdoor scenes (500 of each). We also compared the subjective visual effects on real-world hazy images used in the literature. Ablation studies were conducted on TestSet-S containing 400 hazy indoor/outdoor images, a dataset initially used in a challenge \cite{ChinaMM18dehazing}.

\begin{figure}
\centering
\includegraphics[width=0.83\linewidth]{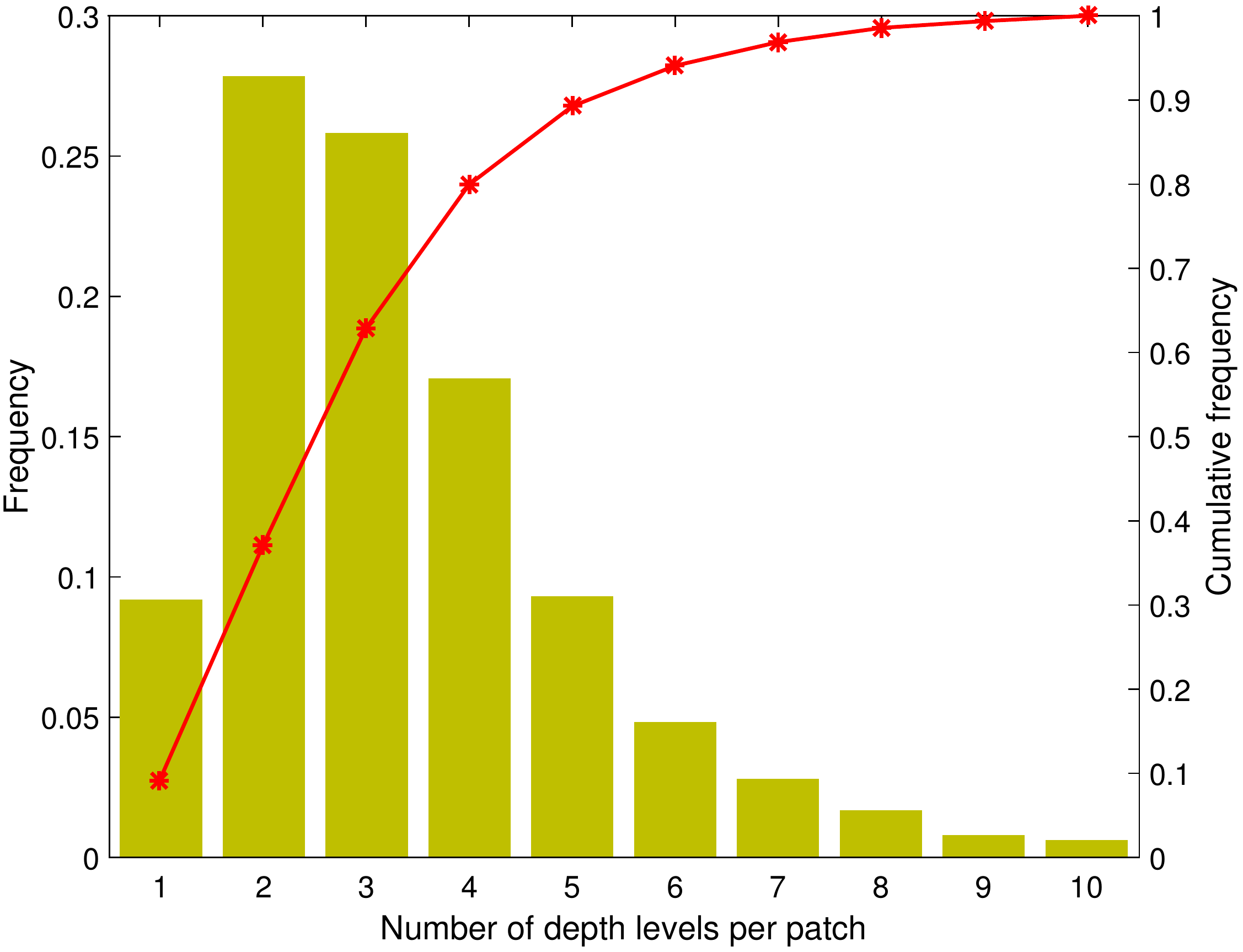}
\caption{Statistics of depth levels of image patches ($128\times128$) in the RESIDE training set. }
\label{fig:depthLevelStats}
\end{figure}

FAMED-Net was trained for a total of 400,000 iterations on the combination of ITS and OTS in RESIDE. 128x128 patches randomly cropped from training images were used for training. Figure~\ref{fig:depthLevelStats} shows the corresponding statistics of depth levels within the training patches. We quantized depth maps into 10 uniform levels according to the maximum and minimum depth values. Then, we counted the number of unique depth levels within each patch and calculated the histogram and its corresponding cumulative distribution as shown in Figure~\ref{fig:depthLevelStats}. As can be seen, almost 65\% patches cover at least 3 depth levels and more than 40\% patches cover at least 4 depth levels. It is noteworthy that since the sizes of training images from different scenes are around $550\times400$, each $128\times128$ patch could cover diverse scene structures as evident by the statistics. Consequently, there are different levels of haze in each patch, $i.e.$, light and dense haze. It facilitates FAMED-Net with a receptive field of $52\times52$ to learn effective feature representation while avoiding overfitting plain structures.

Hyper-parameters were tuned on the validation set. The batch size was set to 48. The initial learning rate was set to 0.00001, which decreased by 10 after 200,000 and 320,000 iterations. The momentum and weight decay were set to 0.9 and 0.0001, respectively. Average pooling was used unless otherwise specified. During testing, the kernel radius of the fast-guided filter was set to 48. The regularization parameter epsilon was set to 0.0001. The down-sampling factor was set to 4. FAMED-Net was implemented in Caffe \cite{jia2014caffe} and run on a workstation with a 3.5 GHz CPU, 32G RAM, and Nvidia Titan XP GPUs.

\subsection{Ablation Experiments}
\label{subsec:exp_ablation}

\begin{table*}[htbp]
  \centering
  \caption{Results of the different basic FAMED-Net architectures on RESIDE TestSet-S. The best and second-best scores are highlighted in red and blue, respectively.}
    \begin{tabular}{lccccccccc}
    \toprule
     & \multicolumn{3}{c}{} & \multicolumn{3}{c}{PSNR (dB)} & \multicolumn{3}{c}{SSIM} \\
     \cmidrule{5-10}
    Model & BN    & Feat. Dim & Scales & Indoor & Outdoor & Average & Indoor & Outdoor & Average \\
    \midrule
    FAMED-Net-NoBN & \text{\sffamily x}   & 4     & 1     & 18.78 & 24.68 & 21.73 & 0.7117 & 0.8949 & 0.8033 \\
    FAMED-Net-FD4 & \checkmark   & 4     & 1     & 20.14 & 25.82 & 22.98 & 0.7488 & 0.9296 & 0.8392 \\
    FAMED-Net-FD8 & \checkmark   & 8     & 1     & 20.45 & 25.88 & 23.17 & 0.7785 & 0.9195 & 0.8490 \\
    FAMED-Net-FD16 & \checkmark   & 16    & 1     & 20.39 & 25.69 & 23.04 & 0.7905 & 0.9306 & 0.8605 \\
    \textbf{FAMED-Net-S} & \checkmark   & 32    & 1     & \textcolor[rgb]{0,0,1}{\textbf{20.71}} & 25.71 & 23.21 & \textcolor[rgb]{0,0,1}{\textbf{0.7958}} & 0.9307 & 0.8633 \\
    FAMED-Net-GP2 & \checkmark   & 32    & 2     & 20.66 & 26.19 & 23.42 & 0.7940 & \textcolor[rgb]{0,0,1}{\textbf{0.9312}} & 0.8626 \\
    \textbf{FAMED-Net-GP} & \checkmark   & 32    & 3     & \textcolor[rgb]{1,0,0}{\textbf{20.85}} & \textcolor[rgb]{0,0,1}{\textbf{26.22}} & \textcolor[rgb]{0,0,1}{\textbf{23.54}} & \textcolor[rgb]{1,0,0}{\textbf{0.8051}} & 0.9268 & \textcolor[rgb]{1,0,0}{\textbf{0.8660}} \\
    FAMED-Net-GP-FD64 & \checkmark   & 64    & 3     & 20.59 & \textcolor[rgb]{1,0,0}{\textbf{26.61}} & \textcolor[rgb]{1,0,0}{\textbf{23.60}} & 0.7929 & \textcolor[rgb]{1,0,0}{\textbf{0.9362}} & \textcolor[rgb]{0,0,1}{\textbf{0.8646}} \\
    \bottomrule
    \end{tabular}%
  \label{tab:ablationArch}%
\end{table*}%

\begin{table*}[htbp]
  \centering
  \caption{Results of FAMED-Net-GP trained with different volumes of training data and iterations on RESIDE TestSet-S.}
    \begin{tabular}{ccccccccc}
    \toprule
          &       &       & \multicolumn{3}{c}{PSNR (dB)} & \multicolumn{3}{c}{SSIM} \\
          \cmidrule{4-9}
    Model & Training Data & Iterations & Indoor & Outdoor & Average & Indoor & Outdoor & Average \\
    \midrule
    FAMED-Net-GP & 40,000 & 100,000 & 20.85 & 26.22 & 23.54 & 0.8051 & 0.9268 & 0.8660 \\
    FAMED-Net-GP & 40,000 & 400,000 & \textcolor[rgb]{0,0,1}{\textbf{20.89}} & \textcolor[rgb]{0,0,1}{\textbf{26.67}} & \textcolor[rgb]{0,0,1}{\textbf{23.78}} & \textcolor[rgb]{0,0,1}{\textbf{0.8082}} & \textcolor[rgb]{0,0,1}{\textbf{0.9357}} & \textcolor[rgb]{0,0,1}{\textbf{0.8719}} \\
    FAMED-Net-GP & ALL(424,450) & 400,000 & \textcolor[rgb]{1,0,0}{\textbf{23.42}} & \textcolor[rgb]{1,0,0}{\textbf{27.94}} & \textcolor[rgb]{1,0,0}{\textbf{25.68}} & \textcolor[rgb]{1,0,0}{\textbf{0.8687}} & \textcolor[rgb]{1,0,0}{\textbf{0.9483}} & \textcolor[rgb]{1,0,0}{\textbf{0.9085}} \\
    \bottomrule
    \end{tabular}%
  \label{tab:ablationTraingsetAndIters}%
\end{table*}%

\begin{table*}[htbp]
  \centering
  \caption{Results of different variants of FAMED-Net-GP on RESIDE TestSet-S.}
    \begin{tabular}{lccccccccc}
    \toprule
          &       &       &       & \multicolumn{3}{c}{PSNR (dB)} & \multicolumn{3}{c}{SSIM}\\
    \cmidrule{5-10}
    Model & 3x3 Conv. & Training Data & Iterations & Indoor & Outdoor & Average & Indoor & Outdoor & Average \\
    \midrule
    FAMED-Net-GP-3x3 & \checkmark(4) & 40,000 & 100,000 & 20.62 & 26.83 & 23.73 & 0.7851 & 0.9427 & 0.8639 \\
    FAMED-Net-GP-3x3 & \checkmark(8) & 40,000 & 100,000 & 21.07 & 26.53 & 23.80 & 0.8189 & 0.9445 & 0.8817 \\
    FAMED-Net-GP-3x3 & \checkmark(8) & ALL(424,450) & 400,000 & \textcolor[rgb]{0,0,1}{\textbf{24.02}} & \textcolor[rgb]{0,0,1}{\textbf{27.86}} & \textcolor[rgb]{0,0,1}{\textbf{25.94}} & \textcolor[rgb]{1,0,0}{\textbf{0.8840}} & \textcolor[rgb]{0,0,1}{\textbf{0.9520}} & \textcolor[rgb]{1,0,0}{\textbf{0.9180}} \\
    FAMED-Net-LP & \text{\sffamily x}    & ALL(424,450) & 400,000 &   23.35    &   27.85   &    25.60   &    0.8724   &    0.9492   & 0.9108 \\
    \textbf{FAMED-Net-GP-MaxP} & \text{\sffamily x}    & ALL(424,450) & 400,000 & \textcolor[rgb]{1,0,0}{\textbf{24.34}} & \textcolor[rgb]{1,0,0}{\textbf{28.67}} & \textcolor[rgb]{1,0,0}{\textbf{26.51}} & \textcolor[rgb]{0,0,1}{\textbf{0.8797}} & \textcolor[rgb]{1,0,0}{\textbf{0.9555}} & \textcolor[rgb]{0,0,1}{\textbf{0.9176}} \\
    \bottomrule
    \end{tabular}%
  \label{tab:MSVariants}%
\end{table*}%

\subsubsection{Ablations on the Basic Architecture}
\label{subsubsec:exp_ablation_basicArc}
First, we conducted ablations on the components of the basic FAMED-Net architecture. We sampled a total of 40,000 images from ITS and OTS evenly to form a training set for ablations. Moreover, the models were trained in a total of 100,000 iterations. The learning rate decreased by 0.1 after 50,000 and 80,000 iterations. All other parameters were as described above. The results on TestSet-S are listed in Table~\ref{tab:ablationArch}.

The dehazing results of FAMED-Net-FD4 with batch normalization were much better than FAMED-Net-NoBN. FAMED-Net-FD4 was also found to converge faster than FAMED-Net-NoBN. We also show the impact of the number of convolutional feature channels on the dehazing results. With more channels, the model tended to have a stronger representational capacity and achieved higher PSNR and SSIM scores. For example, FAMED-Net-S achieved a gain of 0.3 dB and 0.024 SSIM score over FAMED-Net-FD4 and a gain of 1.5 dB and 0.06 SSIM score over FAMED-Net-NoBN. With respect to the multi-scale architecture, with an additional down-scale branch, the PSNR score was improved by 0.2 dB but the SSIM score only decreased marginally. With all three scales, FAMED-Net-GP was the best architecture. Finally, we increased the feature channels in FAMED-Net-GP, but this only marginally improved the PSNR score and decreased the SSIM score. As a trade-off between accuracy and complexity, we chose FAMED-Net-GP as the representative architecture.

\subsubsection{Ablations on Training Data Volume and Training Iterations}
\label{subsubsec:exp_ablation_dataAndIters}

We next investigated the impact of training data volume and training iterations. Specifically, we trained FAMED-Net-GP with 400,000 iterations and all the images in ITS and OTS, $i.e.$, a total of 424,450 images. The results are listed in Table~\ref{tab:ablationTraingsetAndIters}. It can be seen that with sufficient training, FAMED-Net-GP improved. Moreover, the PSNR and SSIM significantly improved when FAMED-Net-GP was trained with all the images, producing a gain of 2.14 dB and 0.0425 SSIM score. Therefore, more training data benefits the deep neural network by exploiting its powerful representation capacity.

\subsection{Variants of the Multi-scale Architecture}
\label{subsec:exp_ablation_MSvariants}

\subsubsection{Additional 3x3 Convolutions for Learning Structural Features}
\label{subsubsec:exp_MS_3x3}
Due to the fully point-wise convolutional structure, FAMED-Net-GP has limited ability on learn structural features. To see whether additional structural features benefit dehazing, we inserted additional 3x3 convolutional layers at the beginning of each scale in FAMED-Net-GP (denoted FAMED-Net-GP-3x3). We tested different feature channel configurations including 4 and 8. The results are shown in the first three rows in Table~\ref{tab:MSVariants}.

Compared with FAMED-Net-GP (see the first and last rows in Table~\ref{tab:ablationTraingsetAndIters}), FAMED-Net-GP-3x3 performed better with the same training settings. With more 3x3 convolutional channels, FAMED-Net-GP-3x3 trained with all training images was the best architecture, $i.e.$, 25.94 dB and 0.9180 SSIM score. Compared with its counterpart without 3x3 convolutional layers, gains of 0.26 dB and 0.01 SSIM score were achieved. However, this came at the cost of additional 6.69\% parameters ($i.e.$, 1152) and 6.66\% FLOPs ($i.e.$, 8.26x10$^6$).

\subsubsection{Laplacian Pyramid Architectures}
\label{subsubsec:exp_MS_RL}
In Section~\ref{subsec:MSNetwork}, we also presented a Laplacian pyramid architecture FAMED-Net-LP (see Figure~\ref{fig:network_multiscale}(b)). Compared with the Gaussian pyramid architecture FAMED-Net-GP (see the last row in Table~\ref{tab:MSVariants}), FAMED-Net-LP achieved a marginally lower PSNR and a marginally higher SSIM. Generally, its performance was comparable to FAMED-Net-GP. Since there was no evident benefit to using residual learning, FAMED-Net-GP was used as our default multi-scale architecture in the following experiments.

\subsubsection{The Effectiveness of Max Pooling}
\label{subsubsec:exp_MS_MaxPooling}
For dehazing, effective local features are usually extracted from \emph{extreme} pixel values including the dark channel (the minimum value of all the channels within a local patch) \cite{he2011single}, local max contrast and saturation \cite{tang2014investigating}, and the learned features using the maxout operation in DehazeNet \cite{cai2016dehazenet}. Inspired by these studies, we hypothesized that \emph{max pooling may be more effective for aggregating local statistics and learning effective features for dehazing}. To verify this hypothesis, we changed the average pooling operations in all the pooling layers to max pooling. This structure is denoted FAMED-Net-GP-MaxP and it was trained using the same settings as FAMED-Net-GP. The results are shown in the last row in Table~\ref{tab:MSVariants}.

Compared with its counterpart using average pooling (last row in Table~\ref{tab:ablationTraingsetAndIters}), FAMED-Net-GP-MaxP achieved a significant gain of 0.83 dB and 0.0091 SSIM score. It also outperformed FAMED-Net-GP-3x3 by 0.57 dB and achieved almost the same SSIM score. Therefore, we chose FAMED-Net-GP-MaxP as the representative model of the proposed architectures due to its light weight (a total of 17,991 parameters) and computational efficiency (1.24x10$^8$ FLOPs). For simplicity, it is denoted FAMED-Net in the following sections.

\subsection{Comparison with State-of-the-art Methods}
\label{subsec:exp_comparisonsSOA}
To evaluate the performance of FAMED-Net, we compared it with several state-of-the-art methods including DCP \cite{CVPR2009_He}, FVR \cite{tarel2009fast}, BCCR \cite{meng2013efficient}, GRM \cite{chen2016robust}, CAP \cite{zhu2015fast}, NLD \cite{berman2016non}, DehazeNet \cite{cai2016dehazenet}, MSCNN \cite{ren2016single}, AOD-Net \cite{li2017all, liu2018improved}, FPCNet \cite{Zhang2018Fully}, GFN \cite{Ren_2018_CVPR} and DCPDN \cite{zhang2018densely}

\subsubsection{Results on RESIDE SOTS}
\label{subsubsec:exp_comparisonsSOA_SOTS}
The PSNR and SSIM scores of the different methods are listed in Table~\ref{tab:SOA_SOTS}. Several observations can be made. 1) CNN-based methods \cite{cai2016dehazenet, Zhang2018Fully, li2017all, liu2018improved, Ren_2018_CVPR} generally outperformed the image prior-based methods \cite{CVPR2009_He, tarel2009fast, meng2013efficient, chen2016robust, zhu2015fast, berman2016non}. By learning features in a data-driven manner, CNN-based dehazing models had stronger representative capacities than image prior-based models, which are usually limited to specific scenarios. 2) CNN architecture matters. For example, FPCNet achieved a significant gain over its counterpart DehazeNet by using a lightweight, fully point-wise convolutional architecture. It achieved the second best SSIM score and even outperformed some complicated networks like AOD-Net, GFN, and DCPDN. Further, by integrating the imaging model into the network architecture, the end-to-end AOD-Net recovered the target haze-free image with higher accuracy than the none end-to-end methods \cite{cai2016dehazenet, ren2016single}. 3) FAMED-Net was the best performing method. Moreover, it significantly improved the PSNR and SSIM scores. For example, FAMED-Net surpassed the second-best methods by a large margin of 3.6 dB and 0.05 SSIM score.

\begin{table}[htbp]
  \centering
  \caption{Results of FAMED-Net and state-of-the-art methods on RESIDE SOTS. Scores in the brackets correspond to the indoor and outdoor subsets, respectively. AOD-Net with an asterisk refers to the fine-tuned model with multi-scale SSIM and L2 loss in \cite{liu2018improved}. The best and second-best scores are highlighted in red and blue, respectively.}
    \begin{tabular}{lll}
    \toprule
     Model  & PSNR (dB) & SSIM \\
    \midrule
    DCP \cite{CVPR2009_He}   & 16.62 & 0.8179 \\
    FVR \cite{tarel2009fast}   & 15.72 & 0.7483 \\
    BCCR \cite{meng2013efficient}  & 16.88 & 0.7913 \\
    GRM \cite{chen2016robust}  & 18.86 & 0.8553 \\
    CAP \cite{zhu2015fast}  & 19.05 & 0.8364 \\
    NLD \cite{berman2016non}  & 17.29 & 0.7489 \\
    \midrule
    DehazeNet \cite{cai2016dehazenet} & 21.14 & 0.8472 \\
    MSCNN \cite{ren2016single} & 17.57 & 0.8102 \\
    FPCNet \cite{Zhang2018Fully} & 21.84 (\textcolor[rgb]{0,0,1}{\textbf{20.92}}/22.75)      & \textcolor[rgb]{0,0,1}{\textbf{0.8872}} (\textcolor[rgb]{0,0,1}{\textbf{0.8729}}/0.9014) \\
    AOD-Net \cite{li2017all} & 19.06 & 0.8504 \\
    AOD-Net* \cite{liu2018improved} & \textcolor[rgb]{0,0,1}{\textbf{23.43}} (20.68/\textcolor[rgb]{0,0,1}{\textbf{26.18}}) & 0.8747 (0.8229/\textcolor[rgb]{0,0,1}{\textbf{0.9266}}) \\
    GFN \cite{Ren_2018_CVPR}   & 22.30  & 0.8800 \\
    DCPDN \cite{zhang2018densely} & 20.81 (19.13/22.49) & 0.8378 (0.8191/0.8565) \\
    \textbf{FAMED-Net} & \textcolor[rgb]{1,0,0}{\textbf{27.01 (25.00/29.03)}} & \textcolor[rgb]{1,0,0}{\textbf{0.9371 (0.9172/0.9570)}} \\
    \bottomrule
    \end{tabular}%
  \label{tab:SOA_SOTS}%
\end{table}%

After carefully dissecting the proposed architecture of FAMED-Net and comparing it with state-of-the-art architectures, we can make the following conclusions. First, point-wise convolution plays a key role in constructing a compact and lightweight dehazing network. Cascaded point-wise convolutional layers are very effective for tackling the ill-posed dehazing problem by aggregating local statistic-based features layer by layer. Second, modeling the dehazing task in an end-to-end manner is beneficial. Third, a carefully designed multi-scale architecture can handle scale variance in complex scenes while only minimally increasing the computational cost. Finally, re-using features via dense connections like \cite{li2017all, zhang2018densely, huang2017densely} leads to a better and more compact model.

\begin{figure*}
\centering
\includegraphics[width=1\linewidth]{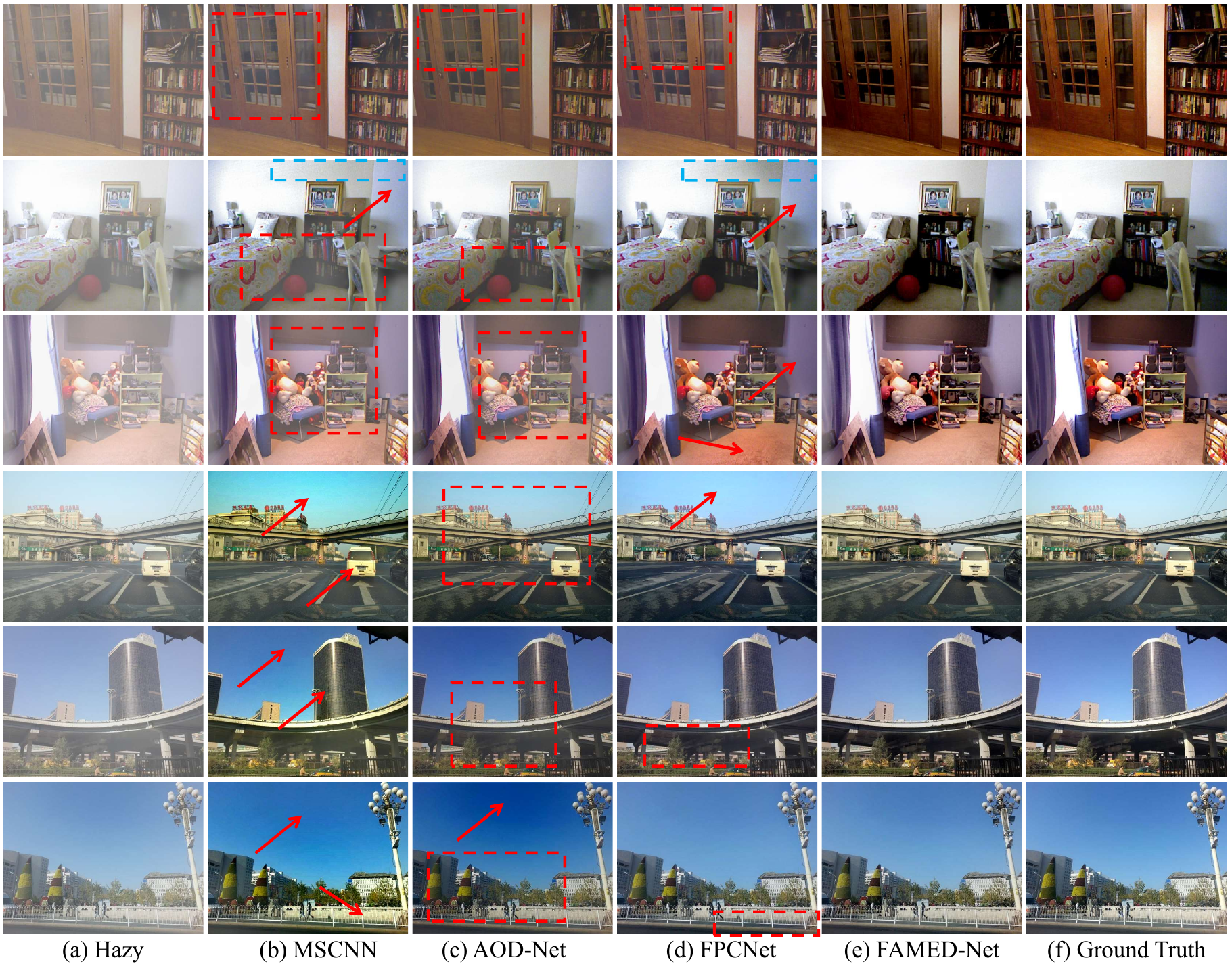}
\caption{Subjective comparisons between FAMED-Net and three most related state-of-the-art methods including MSCNN \cite{ren2016single}, AOD-Net \cite{li2017all}, FPCNet \cite{Zhang2018Fully} on synthetic hazy images from RESIDE test set. Best viewed in color. }
\label{fig:subjectiveEval_syn}
\end{figure*}

\begin{figure*}
\centering
\includegraphics[width=1\linewidth]{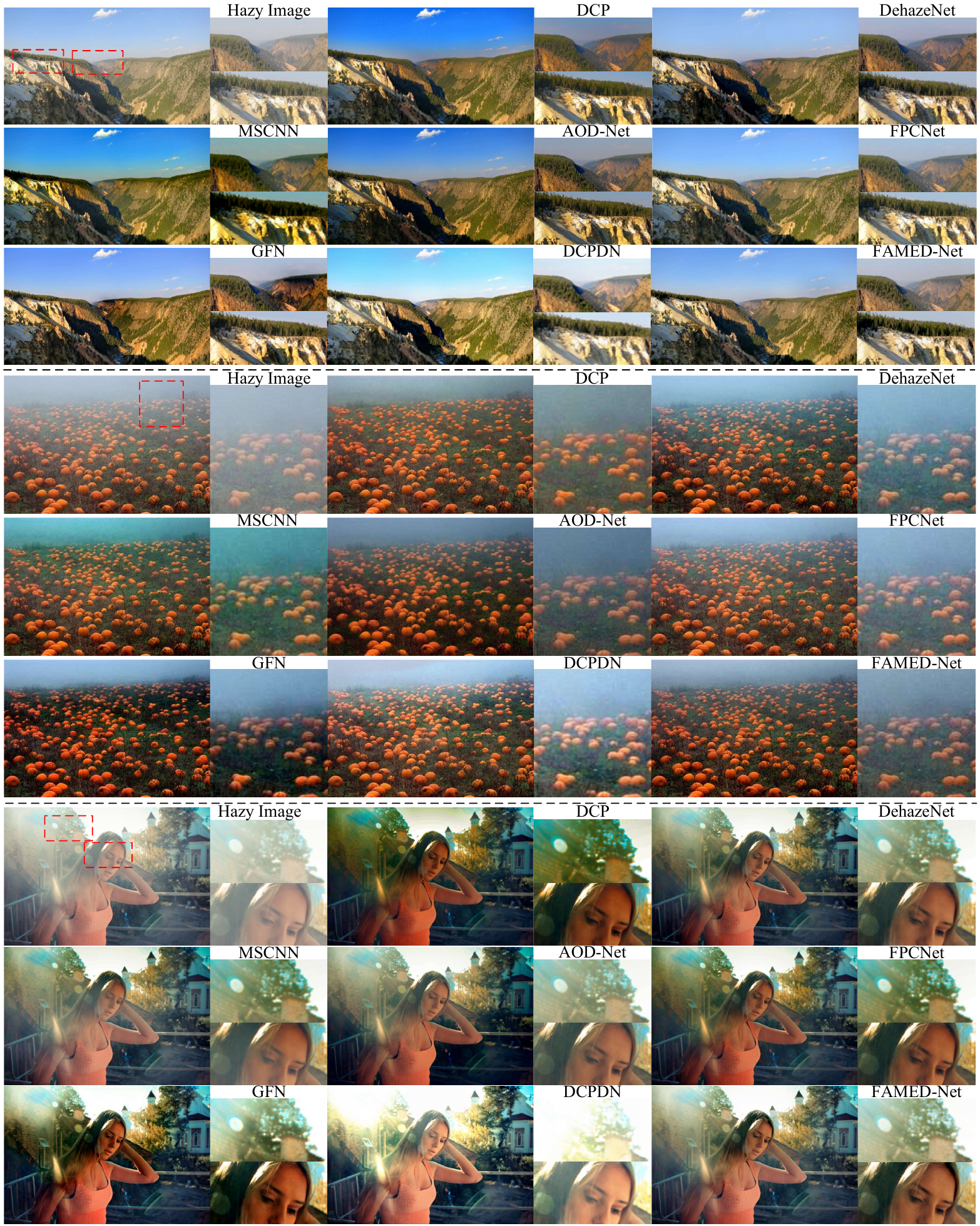}
\caption{Subjective comparisons between FAMED-Net and state-of-the-art methods including DCP \cite{CVPR2009_He}, DehazeNet \cite{cai2016dehazenet}, MSCNN \cite{ren2016single}, AOD-Net \cite{li2017all}, FPCNet \cite{Zhang2018Fully}, GFN \cite{Ren_2018_CVPR} and DCPDN \cite{zhang2018densely} on real-world hazy images. Best viewed in color.}
\label{fig:subjectiveEval}
\end{figure*}

\subsubsection{Subjective Evaluation}
\label{subsubsec:exp_comparisonsSOA_subjective}
Subjective comparison on synthetic hazy images are presented in Figure~\ref{fig:subjectiveEval_syn}. Dehazed results of MSCNN \cite{ren2016single} on indoor images have residual haze indicated by the red boxes. Besides, MSCNN tended to produce over-saturated results with color distortions as indicated by the red arrows. Similar phenomena can also be found in the results of AOD-Net \cite{li2017all}. Although FPCNet \cite{Zhang2018Fully} achieved better results, there are some haze residual and color distortions as well. Moreover, MSCNN and FPCNet produced noisy results due to the incorrectly estimated transmission in regions enclosed by the blue boxes. The proposed FAMED-Net successfully restores the clear images with higher color fidelity and less haze/noise residual. It demonstrates the  fitting ability of FAMED-Net learned from synthetic training images.

Next, we present the results on real-world hazy images in Figure~\ref{fig:subjectiveEval} to compare different methods' generalization ability. Close-up views in the red rectangles are also presented. It can be seen that DCP, MSCNN, and AOD-Net tended to produce over-saturated results, especially in sky regions. MSCNN also exhibits color artifacts, making the dehazed results unrealistic (see the first two images). AOD-Net dehazed images appear dimmer than the others. DehazeNet achieved better results, but still produced some color artifacts (see the middle part of the first image and the bluish artifact in the second image). FPCNet outperformed DehazeNet but retained some haze.

Using some enhanced results as input and a fusion strategy, GFN generated visually better results. However, color distortions in the middle part of the first image and the over-saturated second image are visually unpleasant. DCPDN produced better dehazing results and brighter results. However, some details are missing due to the over-exposure-like artifacts. Generally, FAMED-Net produced better or at least comparable results to state-of-the-art methods, $i.e.$, clear details with fewer color artifacts and high-fidelity sky regions. We also compared image enhancement for anti-halation using different methods in the last row. FAMED-Net also produced visually pleasing results. More results can be found in the supplement.

\subsubsection{Cross-set Generalization}
\label{subsubsec:exp_comparisonsSOA_crosssetEval}
We also compared the cross-set generalization between FAMED-Net and two recently proposed methods, GFN and DCPDN. We used RESIDE SOTS and TestA in \cite{zhang2018densely} as two test sets. We used the pre-trained models of all three methods and did not fine-tune them. The results are listed in Table~\ref{tab:crossSet}. It can be seen that FAMED-Net shows better generalization than GFN and DCPDN, which we ascribe to using the large-scale training set and the effectiveness of the proposed architecture.

\begin{table}[htbp]
  \centering
  \caption{Comparison of cross-set generalization.}
    \begin{tabular}{lllll}
    \toprule
          & \multicolumn{2}{c}{RESIDE SOTS} & \multicolumn{2}{c}{DCPDN-TestA \cite{zhang2018densely}} \\
    \cmidrule{2-5}
    Model & PSNR (dB)  & SSIM  & PSNR (dB)  & SSIM \\
    \midrule
    DCPDN \cite{zhang2018densely} & 20.81     & 0.8378     & \textcolor[rgb]{1,0,0}{\textbf{31.32}}    & \textcolor[rgb]{1,0,0}{\textbf{0.9595}} \\
    GFN \cite{Ren_2018_CVPR}  & \textcolor[rgb]{0,0,1}{\textbf{22.30}}  & \textcolor[rgb]{0,0,1}{\textbf{0.8800}}  & 21.49 & 0.8535 \\
    \textbf{FAMED-Net} & \textcolor[rgb]{1,0,0}{\textbf{27.01}} & \textcolor[rgb]{1,0,0}{\textbf{0.9371}} & \textcolor[rgb]{0,0,1}{\textbf{25.65}} & \textcolor[rgb]{0,0,1}{\textbf{0.9088}} \\
    \bottomrule
    \end{tabular}%
  \label{tab:crossSet}%
\end{table}%

\begin{figure*}
\centering
\includegraphics[width=1\linewidth]{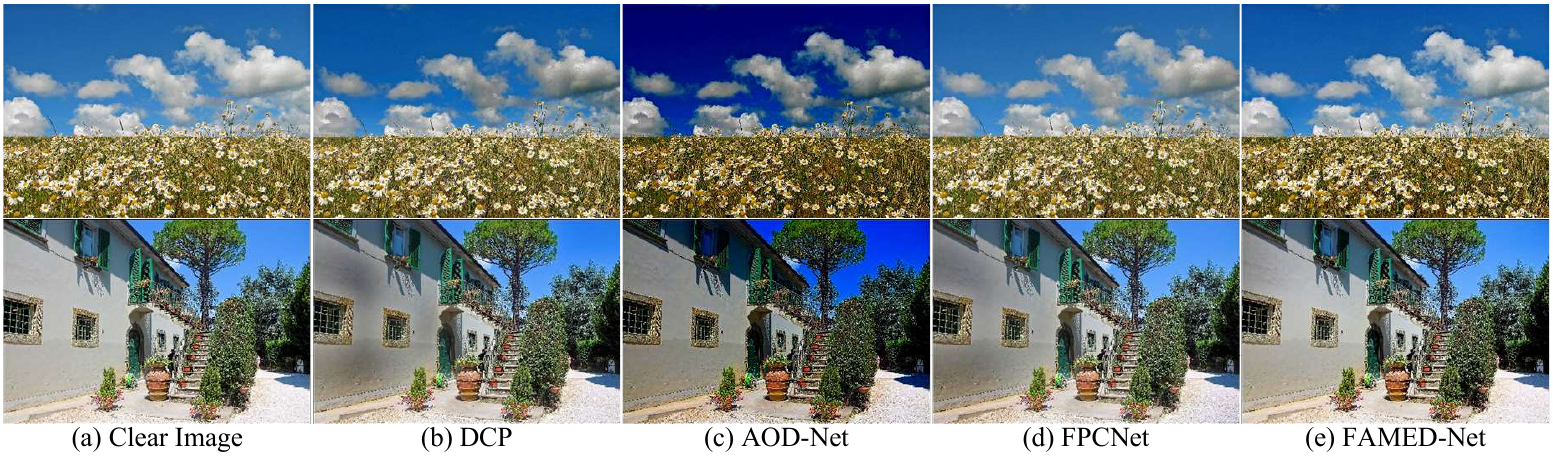}
\caption{Dehazed results of DCP \cite{CVPR2009_He}, AOD-Net \cite{li2017all}, FPCNet \cite{Zhang2018Fully} and FAMED-Net on haze-free images.}
\label{fig:clearDehazed}
\end{figure*}

\subsubsection{Analysis on the Learned Latent Statistical Regularities}
\label{subsubsec:exp_comparisonsStatistical}

Image prior-based methods including DCP \cite{CVPR2009_He}, CAP \cite{zhu2015fast} and NLD \cite{berman2016non} assume prior statistics on haze-free images, which are used to enforce statistical regularities on recovering the target dehazed results \cite{Aapo2009Natural}. The learning-based methods also learn latent statistical regularities \cite{cai2016dehazenet, li2017all, Zhang2018Fully}. For example, DehazeNet and FPCNet, which regress the transmission, should produce a transmission map of all 1s for a haze-free image. In other words, they should learn dark channel-like statistical priors, $i.e.$, $1 - t \approx 0$. As for AOD-Net and FAMED-Net, they regress a latent variable K implicitly. For a haze-free image, the atmospheric light is usually assumed to be white, $i.e.$, $[1,1,1]$. Therefore, the corresponding $K$ can be deduced as $K = \frac{1}{t}$ from Eq.~\eqref{eq:K}. Also, it should be a map all of 1s, $i.e.$, $1 - \frac{1}{{\widehat K}} \approx 0$, where ${\widehat K}$ is the mean across three channels.

\begin{figure}
\centering
\subfloat[]{\label{fig:Stats_Hist}
\includegraphics[height=0.42\linewidth]{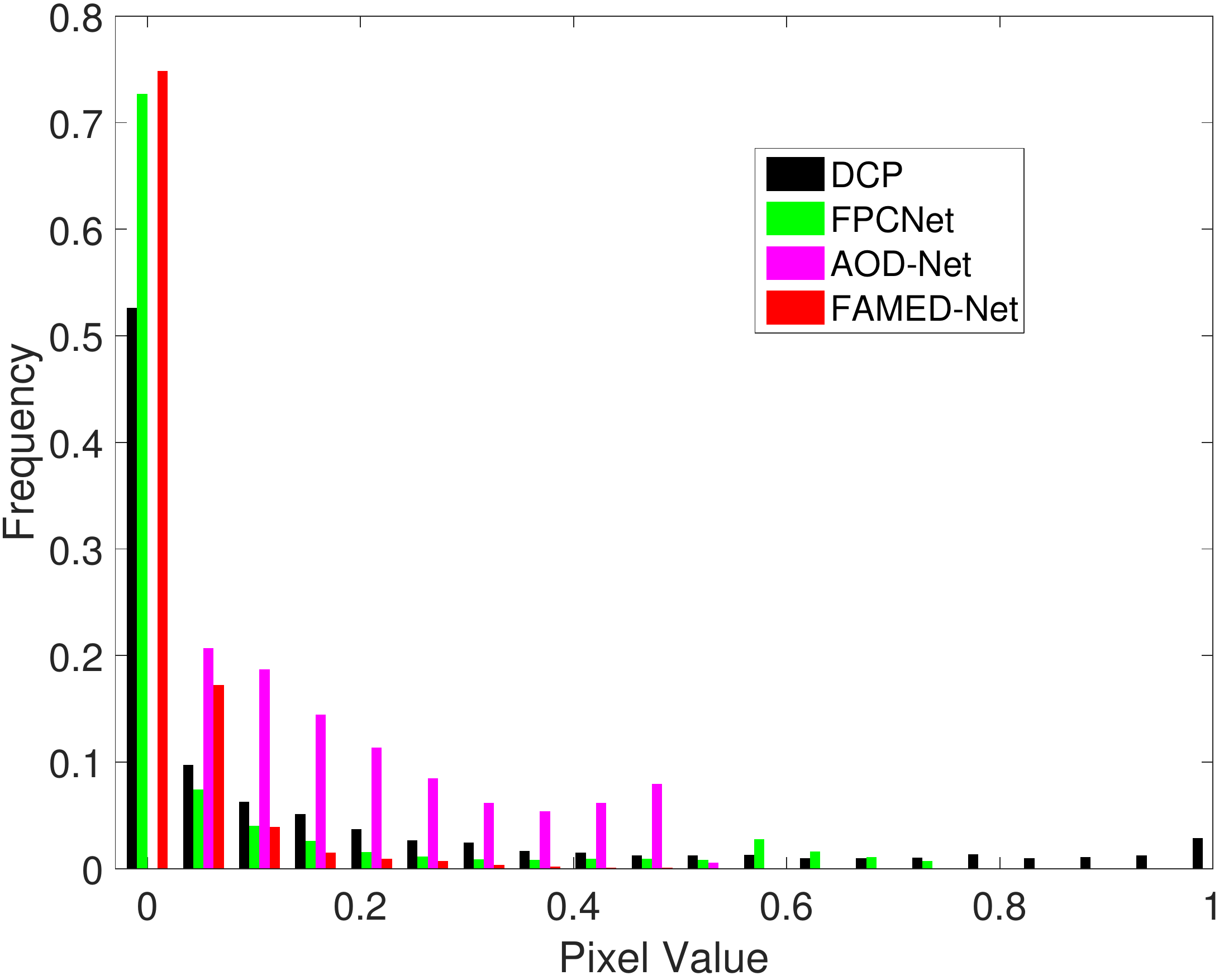}}
\hspace{0.0001\linewidth}
\subfloat[]{\label{fig:Stats_CumulativeHist}
\includegraphics[height=0.42\linewidth]{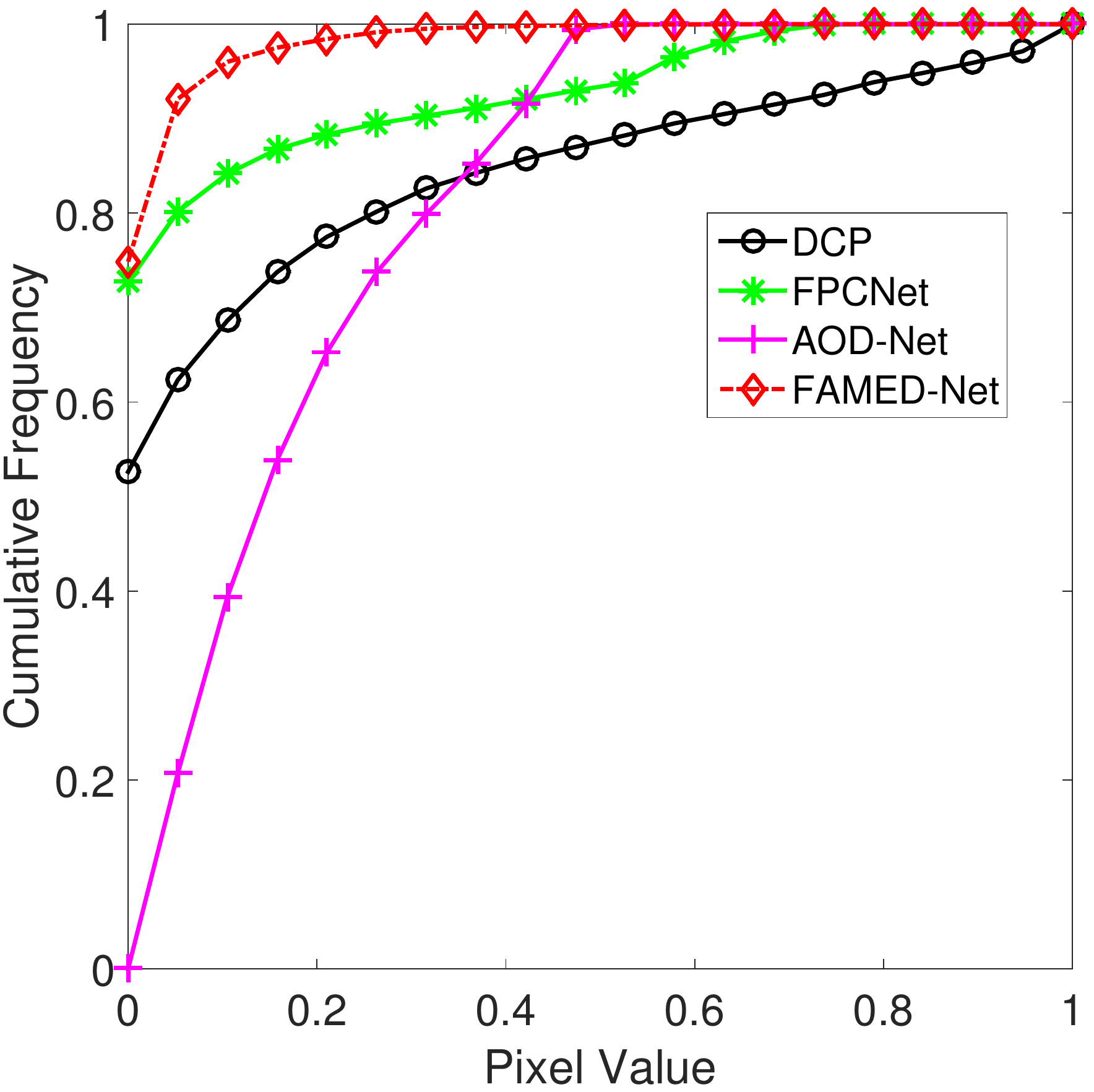}}

\caption{The learned latent statistical regularities of AOD-Net \cite{li2017all}, FPCNet \cite{Zhang2018Fully} and FAMED-Net on haze-free images.}
\label{fig:Learnedhistograms}
\end{figure}

To compare the learned statistical regularities of different methods, we collected 100 haze-free images (two examples are shown in the first column of Figure~\ref{fig:clearDehazed}) . These images were resized such that the long side was 480 pixels and the short side ranged from 100 to 480 pixels. Then, we calculated the dark channel, $t$, and $K$ within each local patch of size $7\times7$. Next, we split the range of pixel value into 20 uniform bin centers and counted the corresponding number of pixels belonging to each bin on all images. Finally, we plotted the histograms of dark channel, $1 - t$, and $1 - \frac{1}{{\widehat K}}$ for DCP, FPCNet, AOD-Net, and FAMED-Net in Figure~\ref{fig:Learnedhistograms}. FAMED-Net learned a much more effective statistical regularity than DCP, FPCNet, and AOD-Net. Besides, the statistics of AOD-Net are far from zero. In other words, the trained network implicitly assumes that there is haze that needs to be removed in haze-free images. Therefore, it leads to over-dehazed artifacts, as seen in the third column. This is consistent with the visual results in Figure~\ref{fig:subjectiveEval}.

\subsubsection{Runtime Analysis}
\label{subsubsec:exp_comparisonsSOA_runningTime}
Following \cite{li2018benchmarking}, we further compared the runtime of different methods on the indoor images ($620\times460$) in RESIDE SOTS. The results are listed in Table~\ref{tab:SOA_runningTime} in Section~\ref{subsec:ComplexityAnalysis}. Results of the classical methods above the line and cGAN are from \cite{li2018benchmarking, Li_2018_CVPR}. Others are reported using our workstation and the code released by the authors. We report the runtime of network forward computation and the whole algorithm including fast-guided filter refinement for FPCNet and FAMED-Net, as shown in separate rows in Table~\ref{tab:SOA_runningTime}. The numbers before/after the slash denote the runtime in CPU/GPU mode, $i.e.$, C/G. FAMED-Net runs very fast and reaches 85 fps and 35 fps without/with fast-guided filter refinement. In addition, we also list the number of parameters and model size of each CNN model. Compared with the recently proposed GFN, cGAN, and DCPDN, FAMED-Net is much more compact and lightweight.

\begin{figure}
\centering
\includegraphics[width=1\linewidth]{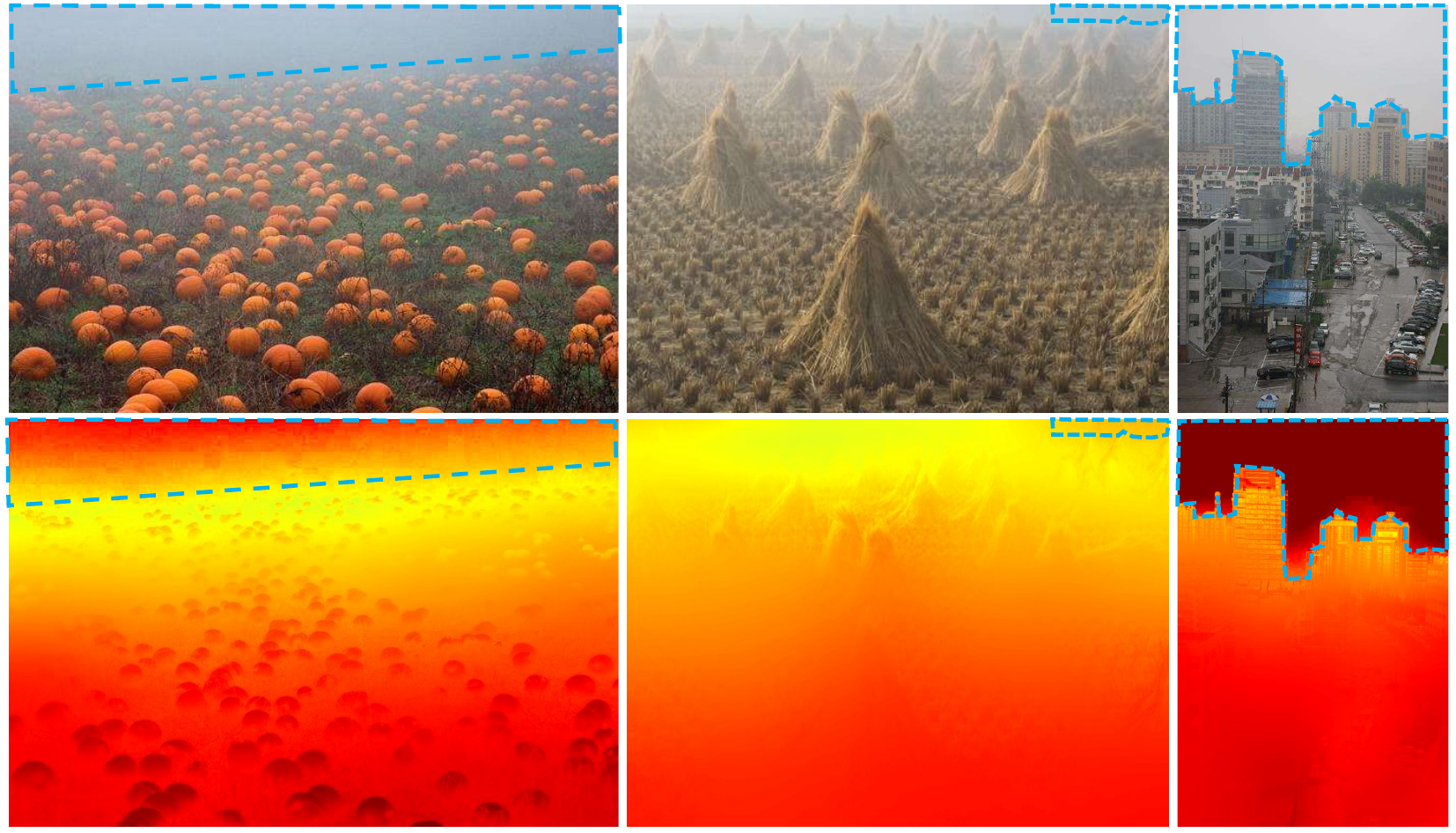}
\caption{Visualization of the estimated transmission maps by FAMED-Net. Warm color (red) represents high transmission, $i.e.$, near camera field with small depth. }
\label{fig:depthVis}
\end{figure}

\subsection{Limitations and Discussions}
\label{subsec:limitation}
As stated in Section \ref{subsec:related_work_dehazing_methods} and demonstrated in Section \ref{subsubsec:exp_comparisonsStatistical}, the proposed FAMED-Net implicitly learns a locally statistical regularity for dehazing like many prior- and learn-based methods \cite{CVPR2009_He, zhu2015fast, cai2016dehazenet, ren2016single, Zhang2018Fully, li2017all}. Though FAMED-Net outperforms these methods by leveraging more efficient architecture, it still has some limitations. Some examples of transmission maps estimated by FAMED-Net are shown in the bottom row in Figure~\ref{fig:depthVis}. As indicated by the blue polygons, the transmission in the sky regions is incorrect, leading to under-dehazed artifacts as shown in Figure~\ref{fig:subjectiveEval}. It may be solved by incorporating high-level semantics into the dehazing network. However, it comes to the ``\emph{chicken and egg}'' dilemma between the low-level enhancement and high-level understanding of degraded images. We suppose that it could be solved by jointly modeling the two correlated problems in a unified framework, which we leave as future work.

Besides, as evident by the low-light enhancement experiments in the supplement and color constancy results in \cite{Zhang2018Fully}, point-wise convolutions could be used for statistical modeling of illumination, color cast, etc. Referring to the haze imaging model in \cite{zhang2017fast}, we will also exploit FAMED-Net's potential for haze removal in the presence of non-uniform atmosphere light, $e.g.$, artificial ambient light in nighttime haze environment. Extending FAMED-Net to remove heterogeneous haze is also promising by investigating region-based techniques, $e.g.$, haze density-aware segmentation.

\section{Conclusions}
\label{sec:conclusions}

In this paper, we introduce a novel fast and accurate multi-scale end-to-end dehazing network called FAMED-Net to tackle the challenging single image dehazing problem. FAMED-Net comprises three encoders at different scales and a fusion module, which is able to efficiently learn the haze-free image directly. Each encoder consists of cascaded point-wise convolutional layers and pooling layers via a densely connected mechanism. By leveraging a fully point-wise structure, FAMED-Net is lightweight and computationally efficient. Extensive experiments on public benchmark datasets and real-world hazy images demonstrate the superiority of FAMED-Net over other top performing models: it is a fast, lightweight, and accurate deep architecture for single image dehazing.


\ifCLASSOPTIONcaptionsoff
  \newpage
\fi



\bibliographystyle{IEEEtran}
\bibliography{TIP2018_Dehazing}

\begin{thebibliography}{10}
\providecommand{\url}[1]{#1}
\csname url@samestyle\endcsname
\providecommand{\newblock}{\relax}
\providecommand{\bibinfo}[2]{#2}
\providecommand{\BIBentrySTDinterwordspacing}{\spaceskip=0pt\relax}
\providecommand{\BIBentryALTinterwordstretchfactor}{4}
\providecommand{\BIBentryALTinterwordspacing}{\spaceskip=\fontdimen2\font plus
\BIBentryALTinterwordstretchfactor\fontdimen3\font minus
  \fontdimen4\font\relax}
\providecommand{\BIBforeignlanguage}[2]{{%
\expandafter\ifx\csname l@#1\endcsname\relax
\typeout{** WARNING: IEEEtran.bst: No hyphenation pattern has been}%
\typeout{** loaded for the language `#1'. Using the pattern for}%
\typeout{** the default language instead.}%
\else
\language=\csname l@#1\endcsname
\fi
#2}}
\providecommand{\BIBdecl}{\relax}
\BIBdecl

\bibitem{li2017end}
B.~Li, X.~Peng, Z.~Wang, J.~Xu, and D.~Feng, ``End-to-end united video dehazing
  and detection,'' \emph{arXiv preprint arXiv:1709.03919}, 2017.

\bibitem{liu2018improved}
Y.~Liu, G.~Zhao, B.~Gong, Y.~Li, R.~Raj, N.~Goel, S.~Kesav, S.~Gottimukkala,
  Z.~Wang, W.~Ren \emph{et~al.}, ``Improved techniques for learning to dehaze
  and beyond: A collective study,'' \emph{arXiv preprint arXiv:1807.00202},
  2018.

\bibitem{li2018benchmarking}
B.~Li, W.~Ren, D.~Fu, D.~Tao, D.~Feng, W.~Zeng, and Z.~Wang, ``Benchmarking
  single image dehazing and beyond,'' \emph{IEEE Transactions on Image
  Processing}, 2018.

\bibitem{tu2005image}
Z.~Tu, X.~Chen, A.~L. Yuille, and S.-C. Zhu, ``Image parsing: Unifying
  segmentation, detection, and recognition,'' \emph{International Journal of
  computer vision}, vol.~63, no.~2, pp. 113--140, 2005.

\bibitem{tarel2010improved}
J.-P. Tarel, N.~Hautiere, A.~Cord, D.~Gruyer, and H.~Halmaoui, ``Improved
  visibility of road scene images under heterogeneous fog,'' in
  \emph{Intelligent Vehicles Symposium (IV), 2010 IEEE}.\hskip 1em plus 0.5em
  minus 0.4em\relax Citeseer, 2010, pp. 478--485.

\bibitem{sakaridis2018semantic}
C.~Sakaridis, D.~Dai, and L.~Van~Gool, ``Semantic foggy scene understanding
  with synthetic data,'' \emph{International Journal of Computer Vision}, pp.
  1--20, 2018.

\bibitem{tan2000enhancement}
K.~Tan and J.~P. Oakley, ``Enhancement of color images in poor visibility
  conditions.'' in \emph{ICIP}, vol.~2, 2000.

\bibitem{schechner2001instant}
Y.~Y. Schechner, S.~G. Narasimhan, and S.~K. Nayar, ``Instant dehazing of
  images using polarization,'' in \emph{Proc. Computer Vision \& Pattern
  Recognition Vol}, vol.~1, 2001, pp. 325--332.

\bibitem{nayar1999vision}
S.~K. Nayar and S.~G. Narasimhan, ``Vision in bad weather,'' in \emph{The IEEE
  International Conference on Computer Vision}, vol.~2.\hskip 1em plus 0.5em
  minus 0.4em\relax IEEE, 1999, pp. 820--827.

\bibitem{liu2018single}
Q.~Liu, X.~Gao, L.~He, and W.~Lu, ``Single image dehazing with depth-aware
  non-local total variation regularization,'' \emph{IEEE Transactions on Image
  Processing}, vol.~27, no.~10, pp. 5178--5191, 2018.

\bibitem{wang2018aipnet}
A.~Wang, W.~Wang, J.~Liu, and N.~Gu, ``Aipnet: Image-to-image single image
  dehazing with atmospheric illumination prior,'' \emph{IEEE Transactions on
  Image Processing}, 2018.

\bibitem{li2018single}
Z.~Li and J.~Zheng, ``Single image de-hazing using globally guided image
  filtering,'' \emph{IEEE Transactions on Image Processing}, vol.~27, no.~1,
  pp. 442--450, 2018.

\bibitem{fattal2008single}
R.~Fattal, ``Single image dehazing,'' \emph{ACM transactions on graphics
  (TOG)}, vol.~27, no.~3, p.~72, 2008.

\bibitem{he2011single}
K.~He, J.~Sun, and X.~Tang, ``Single image haze removal using dark channel
  prior,'' \emph{IEEE transactions on pattern analysis and machine
  intelligence}, vol.~33, no.~12, pp. 2341--2353, 2011.

\bibitem{tang2014investigating}
K.~Tang, J.~Yang, and J.~Wang, ``Investigating haze-relevant features in a
  learning framework for image dehazing,'' in \emph{The IEEE Conference on
  Computer Vision and Pattern Recognition}, 2014, pp. 2995--3000.

\bibitem{zhu2015fast}
Q.~Zhu, J.~Mai, and L.~Shao, ``A fast single image haze removal algorithm using
  color attenuation prior,'' \emph{IEEE Transactions on Image Processing},
  vol.~24, no.~11, pp. 3522--3533, 2015.

\bibitem{berman2016non}
D.~Berman, S.~Avidan \emph{et~al.}, ``Non-local image dehazing,'' in \emph{The
  IEEE conference on computer vision and pattern recognition}, 2016, pp.
  1674--1682.

\bibitem{cai2016dehazenet}
B.~Cai, X.~Xu, K.~Jia, C.~Qing, and D.~Tao, ``Dehazenet: An end-to-end system
  for single image haze removal,'' \emph{IEEE Transactions on Image
  Processing}, vol.~25, no.~11, pp. 5187--5198, 2016.

\bibitem{ren2016single}
W.~Ren, S.~Liu, H.~Zhang, J.~Pan, X.~Cao, and M.-H. Yang, ``Single image
  dehazing via multi-scale convolutional neural networks,'' in \emph{European
  Conference on Computer Vision}.\hskip 1em plus 0.5em minus 0.4em\relax
  Springer, 2016, pp. 154--169.

\bibitem{li2017all}
B.~Li, X.~Peng, Z.~Wang, J.~Xu, and D.~Feng, ``Aod-net: All-in-one dehazing
  network,'' in \emph{The IEEE International Conference on Computer Vision},
  vol.~1, no.~4, 2017, p.~7.

\bibitem{Zhang2018Fully}
J.~Zhang, Y.~Cao, Y.~Wang, C.~Wen, and C.~W. Chen., ``Fully point-wise
  convolutional neural network for modeling statistical regularities in natural
  images,'' in \emph{ACM Multimedia Conference}, 2018.

\bibitem{Ren_2018_CVPR}
W.~Ren, L.~Ma, J.~Zhang, J.~Pan, X.~Cao, W.~Liu, and M.-H. Yang, ``Gated fusion
  network for single image dehazing,'' in \emph{The IEEE Conference on Computer
  Vision and Pattern Recognition (CVPR)}, June 2018.

\bibitem{zhang2018densely}
H.~Zhang and V.~M. Patel, ``Densely connected pyramid dehazing network,'' in
  \emph{The IEEE Conference on Computer Vision and Pattern Recognition (CVPR)},
  2018.

\bibitem{Li_2018_CVPR}
R.~Li, J.~Pan, Z.~Li, and J.~Tang, ``Single image dehazing via conditional
  generative adversarial network,'' in \emph{The IEEE Conference on Computer
  Vision and Pattern Recognition (CVPR)}, June 2018.

\bibitem{bui2018single}
T.~M. Bui and W.~Kim, ``Single image dehazing using color ellipsoid prior,''
  \emph{IEEE Transactions on Image Processing}, vol.~27, no.~2, pp. 999--1009,
  2018.

\bibitem{CVPR2009_He}
K.~He, J.~Sun, and X.~Tang, ``Single image haze removal using dark channel
  prior,'' in \emph{The IEEE conference on computer vision and pattern
  recognition}, 2009.

\bibitem{goodfellow2014generative}
I.~Goodfellow, J.~Pouget-Abadie, M.~Mirza, B.~Xu, D.~Warde-Farley, S.~Ozair,
  A.~Courville, and Y.~Bengio, ``Generative adversarial nets,'' in
  \emph{Advances in neural information processing systems}, 2014, pp.
  2672--2680.

\bibitem{mccartney1976optics}
E.~J. McCartney, ``Optics of the atmosphere: scattering by molecules and
  particles,'' \emph{New York, John Wiley and Sons, Inc., 1976. 421 p.}, 1976.

\bibitem{narasimhan2002vision}
S.~G. Narasimhan and S.~K. Nayar, ``Vision and the atmosphere,''
  \emph{International Journal of Computer Vision}, vol.~48, no.~3, pp.
  233--254, 2002.

\bibitem{Yang_2018_ECCV}
D.~Yang and J.~Sun, ``Proximal dehaze-net: A prior learning-based deep network
  for single image dehazing,'' in \emph{The European Conference on Computer
  Vision (ECCV)}, September 2018.

\bibitem{daubechies1990wavelet}
I.~Daubechies, ``The wavelet transform, time-frequency localization and signal
  analysis,'' \emph{IEEE transactions on information theory}, vol.~36, no.~5,
  pp. 961--1005, 1990.

\bibitem{lowe2004distinctive}
D.~G. Lowe, ``Distinctive image features from scale-invariant keypoints,''
  \emph{International journal of computer vision}, vol.~60, no.~2, pp. 91--110,
  2004.

\bibitem{lai2017deep}
W.-S. Lai, J.-B. Huang, N.~Ahuja, and M.-H. Yang, ``Deep laplacian pyramid
  networks for fast and accurate super-resolution,'' in \emph{The IEEE
  conference on computer vision and pattern recognition}, 2017, pp. 624--632.

\bibitem{yu2018deepexposure}
R.~Yu, W.~Liu, Y.~Zhang, Z.~Qu, D.~Zhao, and B.~Zhang, ``Deepexposure: Learning
  to expose photos with asynchronously reinforced adversarial learning,'' in
  \emph{Advances in Neural Information Processing Systems}, 2018, pp.
  2149--2159.

\bibitem{denton2015deep}
E.~L. Denton, S.~Chintala, R.~Fergus \emph{et~al.}, ``Deep generative image
  models using a laplacian pyramid of adversarial networks,'' in \emph{Advances
  in neural information processing systems}, 2015, pp. 1486--1494.

\bibitem{xu2018lapran}
K.~Xu, Z.~Zhang, and F.~Ren, ``Lapran: A scalable laplacian pyramid
  reconstructive adversarial network for flexible compressive sensing
  reconstruction,'' in \emph{The European Conference on Computer Vision
  (ECCV)}, 2018, pp. 485--500.

\bibitem{chen2018deeplab}
L.-C. Chen, G.~Papandreou, I.~Kokkinos, K.~Murphy, and A.~L. Yuille, ``Deeplab:
  Semantic image segmentation with deep convolutional nets, atrous convolution,
  and fully connected crfs,'' \emph{IEEE transactions on pattern analysis and
  machine intelligence}, vol.~40, no.~4, pp. 834--848, 2018.

\bibitem{Lin_2017_CVPR}
T.-Y. Lin, P.~Dollar, R.~Girshick, K.~He, B.~Hariharan, and S.~Belongie,
  ``Feature pyramid networks for object detection,'' in \emph{The IEEE
  Conference on Computer Vision and Pattern Recognition (CVPR)}, July 2017.

\bibitem{xie2015holistically}
S.~Xie and Z.~Tu, ``Holistically-nested edge detection,'' in \emph{The IEEE
  international conference on computer vision}, 2015, pp. 1395--1403.

\bibitem{lee2015deeply}
C.-Y. Lee, S.~Xie, P.~Gallagher, Z.~Zhang, and Z.~Tu, ``Deeply-supervised
  nets,'' in \emph{Artificial Intelligence and Statistics}, 2015, pp. 562--570.

\bibitem{Aapo2009Natural}
A.~Hyv{\"a}rinen, J.~Hurri, and P.~O. Hoyer, \emph{Natural Image
  Statistics}.\hskip 1em plus 0.5em minus 0.4em\relax Springer-Verlag London,
  2009.

\bibitem{he2013guided}
K.~He, J.~Sun, and X.~Tang, ``Guided image filtering,'' \emph{IEEE transactions
  on pattern analysis and machine intelligence}, vol.~35, no.~6, pp.
  1397--1409, 2013.

\bibitem{zhang2017fast}
J.~Zhang, Y.~Cao, S.~Fang, Y.~Kang, and C.~W. Chen, ``Fast haze removal for
  nighttime image using maximum reflectance prior,'' in \emph{The IEEE
  Conference on Computer Vision and Pattern Recognition}, 2017, pp. 7418--7426.

\bibitem{he2015fast}
K.~He and J.~Sun, ``Fast guided filter,'' \emph{arXiv preprint
  arXiv:1505.00996}, 2015.

\bibitem{tarel2009fast}
J.-P. Tarel and N.~Hautiere, ``Fast visibility restoration from a single color
  or gray level image,'' in \emph{2009 IEEE 12th International Conference on
  Computer Vision}.\hskip 1em plus 0.5em minus 0.4em\relax IEEE, 2009, pp.
  2201--2208.

\bibitem{meng2013efficient}
G.~Meng, Y.~Wang, J.~Duan, S.~Xiang, and C.~Pan, ``Efficient image dehazing
  with boundary constraint and contextual regularization,'' in \emph{The IEEE
  international conference on computer vision}, 2013, pp. 617--624.

\bibitem{chen2016robust}
C.~Chen, M.~N. Do, and J.~Wang, ``Robust image and video dehazing with visual
  artifact suppression via gradient residual minimization,'' in \emph{European
  Conference on Computer Vision}, 2016, pp. 576--591.

\bibitem{ChinaMM18dehazing}
W.~Ren, Z.~Wang, Y.~Guo, g.~Meng, X.~Fan, and J.~Guo, ``Chinamm18dehazing,''
  \emph{https://rwenqi.github.io/ChinaMM18dehazing/}, 2018.

\bibitem{jia2014caffe}
Y.~Jia, E.~Shelhamer, J.~Donahue, S.~Karayev, J.~Long, R.~Girshick,
  S.~Guadarrama, and T.~Darrell, ``Caffe: Convolutional architecture for fast
  feature embedding,'' in \emph{The 22nd ACM international conference on
  Multimedia}.\hskip 1em plus 0.5em minus 0.4em\relax ACM, 2014, pp. 675--678.

\bibitem{huang2017densely}
G.~Huang, Z.~Liu, L.~Van Der~Maaten, and K.~Q. Weinberger, ``Densely connected
  convolutional networks.'' in \emph{CVPR}, vol.~1, no.~2, 2017, p.~3.

\bibitem{land1977retinex}
E.~H. Land, ``The retinex theory of color vision,'' \emph{Scientific american},
  vol. 237, no.~6, pp. 108--129, 1977.

\bibitem{guo2017lime}
X.~Guo, Y.~Li, and H.~Ling, ``Lime: Low-light image enhancement via
  illumination map estimation,'' \emph{IEEE Transactions on Image Processing},
  vol.~26, no.~2, pp. 982--993, 2017.

\bibitem{Chen2018Retinex}
C.~Wei, W.~Wang, W.~Yang, and J.~Liu, ``Deep retinex decomposition for
  low-light enhancement,'' in \emph{British Machine Vision Conference}, 2018.

\bibitem{elad2005retinex}
M.~Elad, ``Retinex by two bilateral filters,'' in \emph{International
  Conference on Scale-Space Theories in Computer Vision}.\hskip 1em plus 0.5em
  minus 0.4em\relax Springer, 2005, pp. 217--229.

\end{thebibliography}
\clearpage

\begin{figure}[t]
\centering
\subfloat[]{\label{fig:lowlight_gt}
\includegraphics[width=0.45\linewidth]{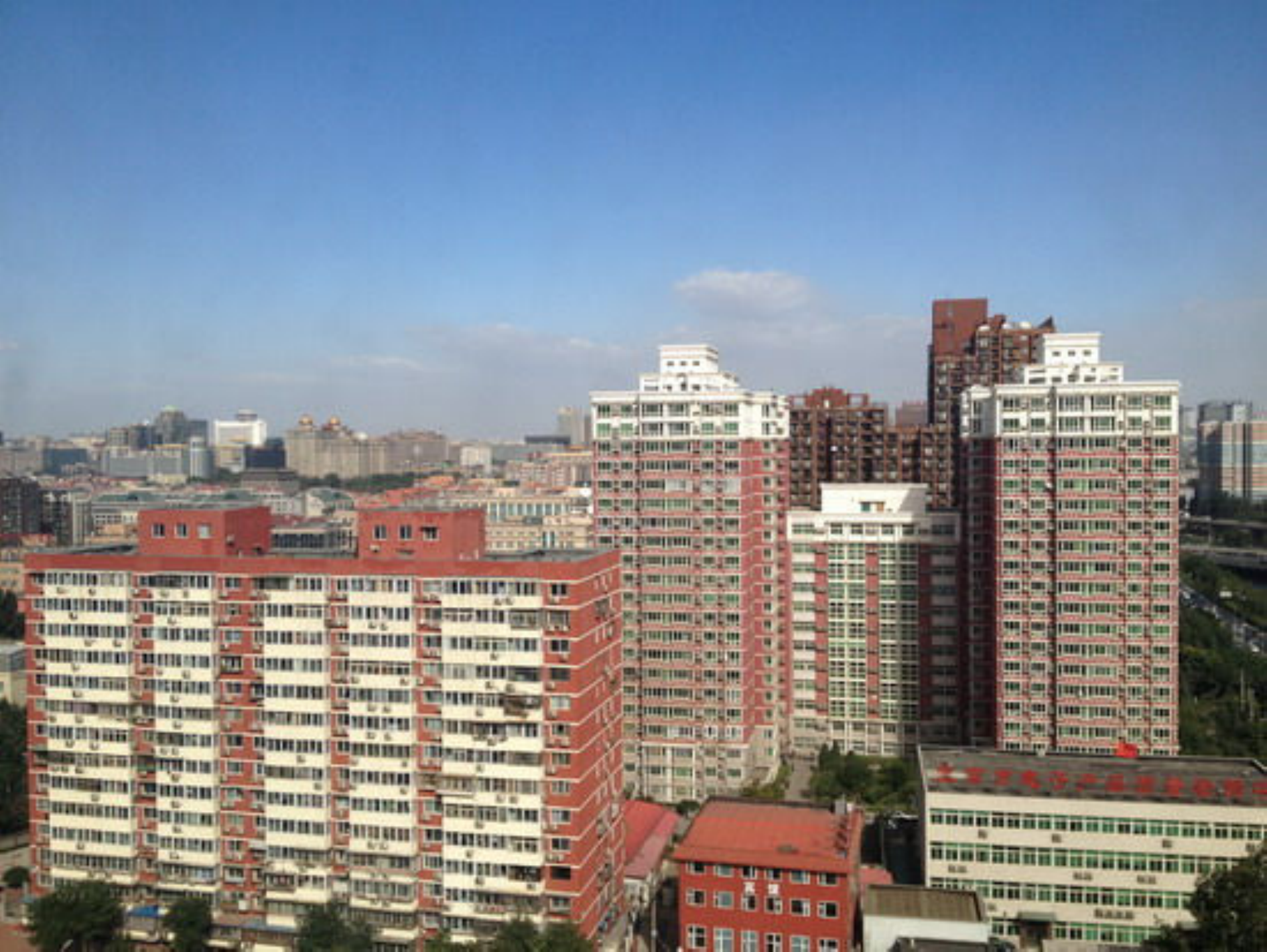}}
\hspace{0.00001\linewidth}
\subfloat[]{\label{fig:lowlight_lowlihgt}
\includegraphics[width=0.45\linewidth]{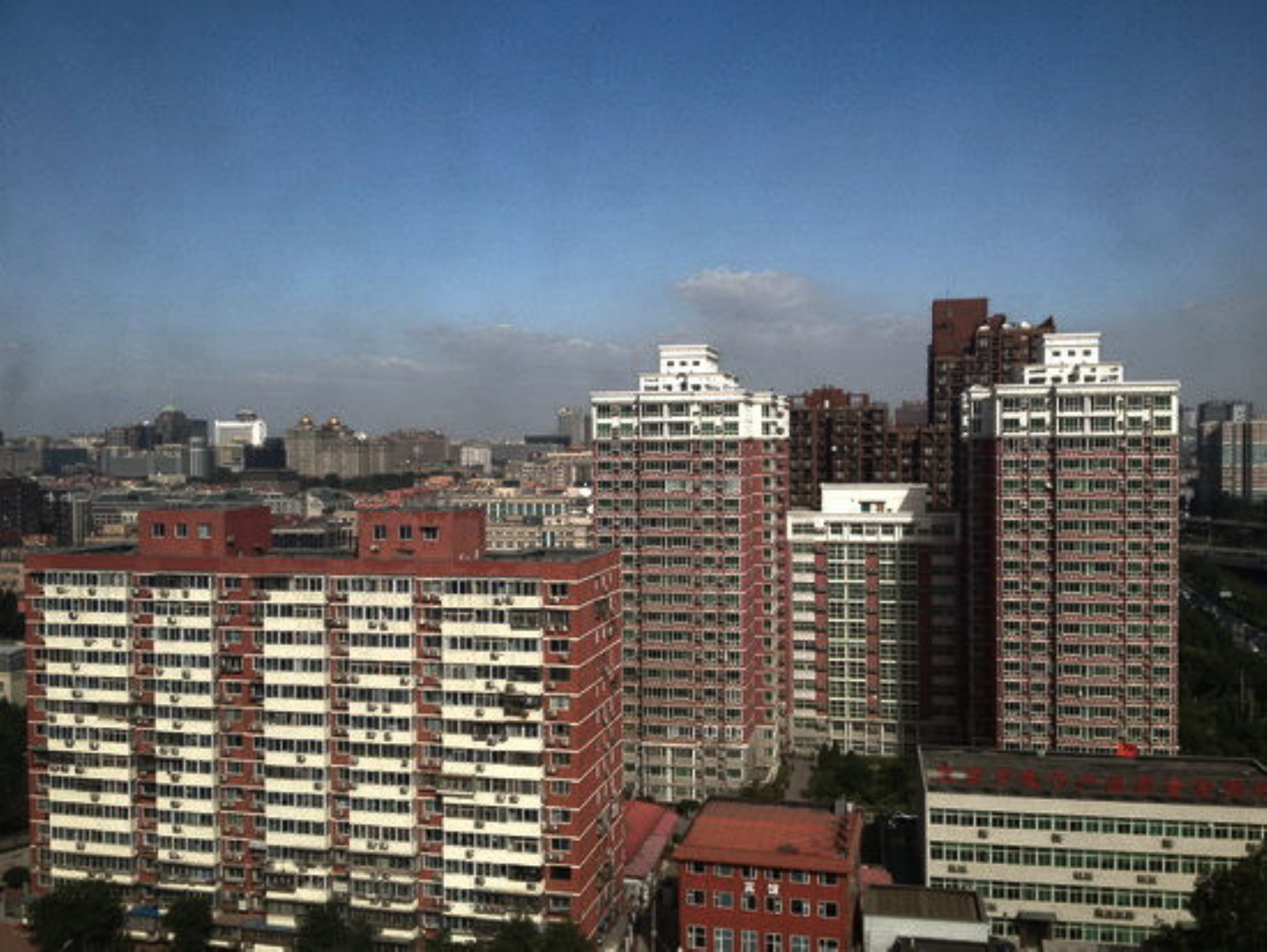}}
\hspace{0.00001\linewidth}
\subfloat[]{\label{fig:lowlight_mapping}
\includegraphics[width=0.45\linewidth]{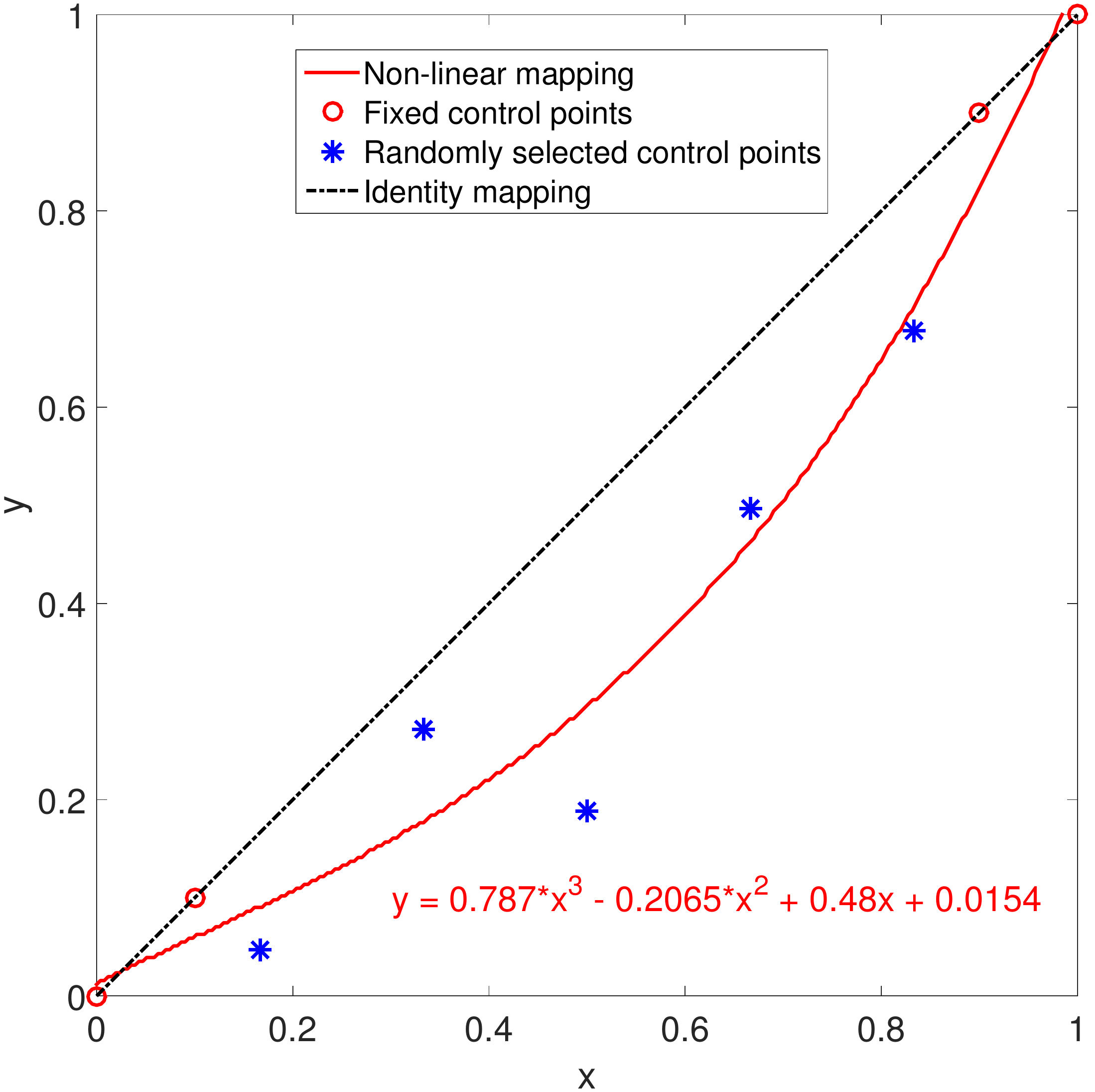}}

\caption{Illustration of generating illumination unbalanced image. (a) A clear image in RESIDE \cite{li2018benchmarking}. (b) The generated illumination unbalanced image according to the non-linear mapping in (c). (c) An exemplar non-linear mapping fitted from the randomly selected control points.}
\label{fig:lowlightNonlinearMapping}
\end{figure}

\section{FAMED-Net: A Fast, Light-weight and Accurate Multi-scale End-to-end Dehazing Network (Supplementary Material)}

\subsection{Illumination Balancing Network}
\label{sec:IlluminationNet}

\subsubsection{Modification of FAMED-Net for Illumination Balancing}
\label{subsec:ModiFAMEDNet}

Since the scene radiance is usually not as bright as the atmospheric light, the recovered haze-free image looks dim \cite{CVPR2009_He}, especially for the dense haze regions and shading regions. It's better to balance the illumination for both visually pleasing and facilitating subsequent high-level tasks. Considering the following imaging model used in Retinex literatures \cite{land1977retinex, guo2017lime, Chen2018Retinex}:
\begin{equation}
{S^\lambda } = {R^\lambda } \circ L,\lambda  \in \left\{ {r,g,b} \right\},
\label{eq:RetinexModel}
\end{equation}
where $S$ represents the observed image, the reflectance $R$ represents the intrinsic property of captured objects, the illumination $L$ represents the various lightness on objects, and $\circ$ denotes element-wise multiplication. Given an observed $S$, estimating $R$ and $L$ is ill-posed. Various smoothness constraints have been proposed to make it tractable \cite{elad2005retinex, guo2017lime, Chen2018Retinex}. Instead of estimating the reflectance which typically looks unrealistic, we follow \cite{elad2005retinex} by retaining some amount of illumination to make it enjoys both the desired brightness and the natural appearance. To this end, we propose a illumination balancing network (IBNet) to estimate a balanced illumination map from an input image. Then we replace the original unbalanced distributed illumination (approximated by the illumination channel in HSV color space) with the estimate. Specifically, we construct the IBNet from FAMED-Net with minor modification: 1) changing the 3-channel $K$ in FAMED-Net to the one-channel illumination map; 2) omitting the recovery module depicted by the yellow circle. We used L2 loss to supervise the estimated illumination map.

To prepare the training/test datasets, we applied a fitted non-linear mapping on the illumination channel of each clear image in RESIDE dataset (See Figure~\ref{fig:lowlightNonlinearMapping}(a)) and used it to replace the original one to form the illumination unbalanced image (See Figure~\ref{fig:lowlightNonlinearMapping}(b)). The non-linear mapping was generated for each image specifically by fitting a cubic curve from some randomly selected control points in the right-bottom half plane as shown in Figure~\ref{fig:lowlightNonlinearMapping}(c) and other four fixed control points, i.e., (0,0), (0.1,0.1), (0.9,0.9) and (1,1).

\begin{table}[htbp]
  \centering
  \caption{PSNR and SSIM scores of IBNet for illumination balancing on RESIDE TestSet-S generated according to Section~\ref{subsec:ModiFAMEDNet}.}
    \begin{tabular}{clcc}
    \toprule
          &       &   Original    & IBNet (FAMED-Net) \\
    \midrule
    \multicolumn{1}{c}{\multirow{3}[0]{*}{PSNR (dB)}} & Indoor & 15.55     & 28.44 \\
    \multicolumn{1}{c}{} & Outdoor & 16.13    & 26.10 \\
    \multicolumn{1}{c}{} & Average & 15.84     & 27.27 \\
    \multicolumn{1}{c}{\multirow{3}[0]{*}{SSIM}} & Indoor & 0.7021    & 0.9316 \\
    \multicolumn{1}{c}{} & Outdoor & 0.7526     & 0.8959 \\
    \multicolumn{1}{c}{} & Average & 0.7273     & 0.9137 \\
    \bottomrule
    \end{tabular}%
  \label{tab:IBNet}%
\end{table}%

\subsubsection{Experimental Results}
\label{subsec:experiments}

\begin{figure*}
\centering
\includegraphics[width=1\linewidth]{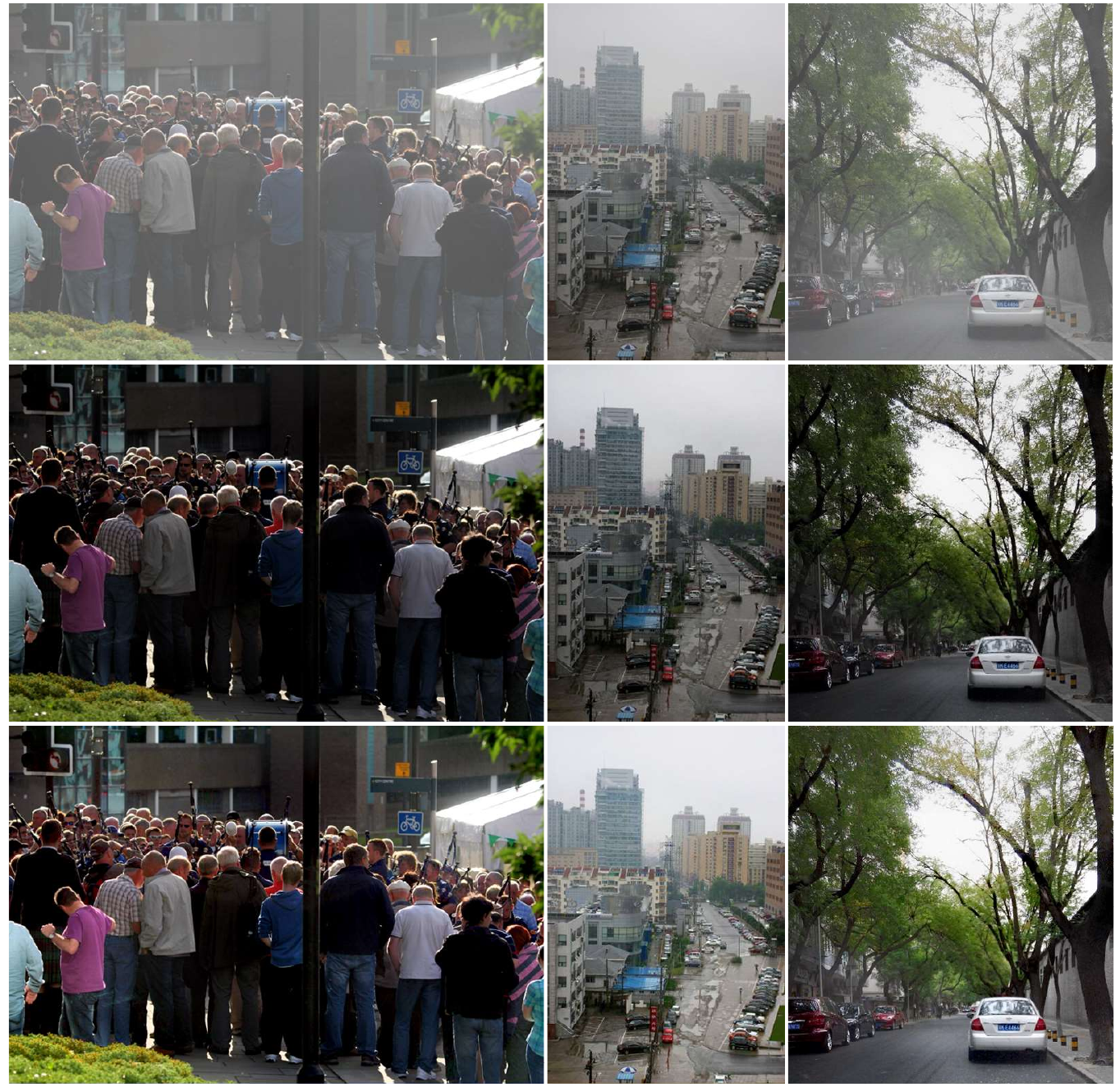}
\caption{Subjective visual inspection on the enhanced results of IBNet. Hazy images, dehazed results by FAMED-Net and illumination balanced results by IBNet are shown from the top row to the bottom row, respectively. Best viewed in color.}
\label{fig:lowlightEnh}
\end{figure*}

We evaluated the proposed IBNet for illumination balancing on RESIDE TestSet-S generated according to Section~\ref{subsec:ModiFAMEDNet}. The results are listed in Table~\ref{tab:IBNet}. As can be seen, IBNet, an incarnation of FAMED-Net, achieved good restoration accuracy by enhancing the unbalanced distributed illumination. Some subjective visual inspection examples are shown in Figure~\ref{fig:lowlightEnh}. As can be seen, the enhancement results of IBNet on the dehazed images are more visually pleasing, e.g., the illumination has been balanced and details are revealed. However, the results also exhibits a few amount of color distortions constrained by the unrealistic synthetic mappings. In future work, we will collect real-world low-light dataset for training a better model.

\subsection{More Subjective Comparisons}
\label{sec:MoreResults}

\begin{figure*}
\centering
\includegraphics[width=1\linewidth]{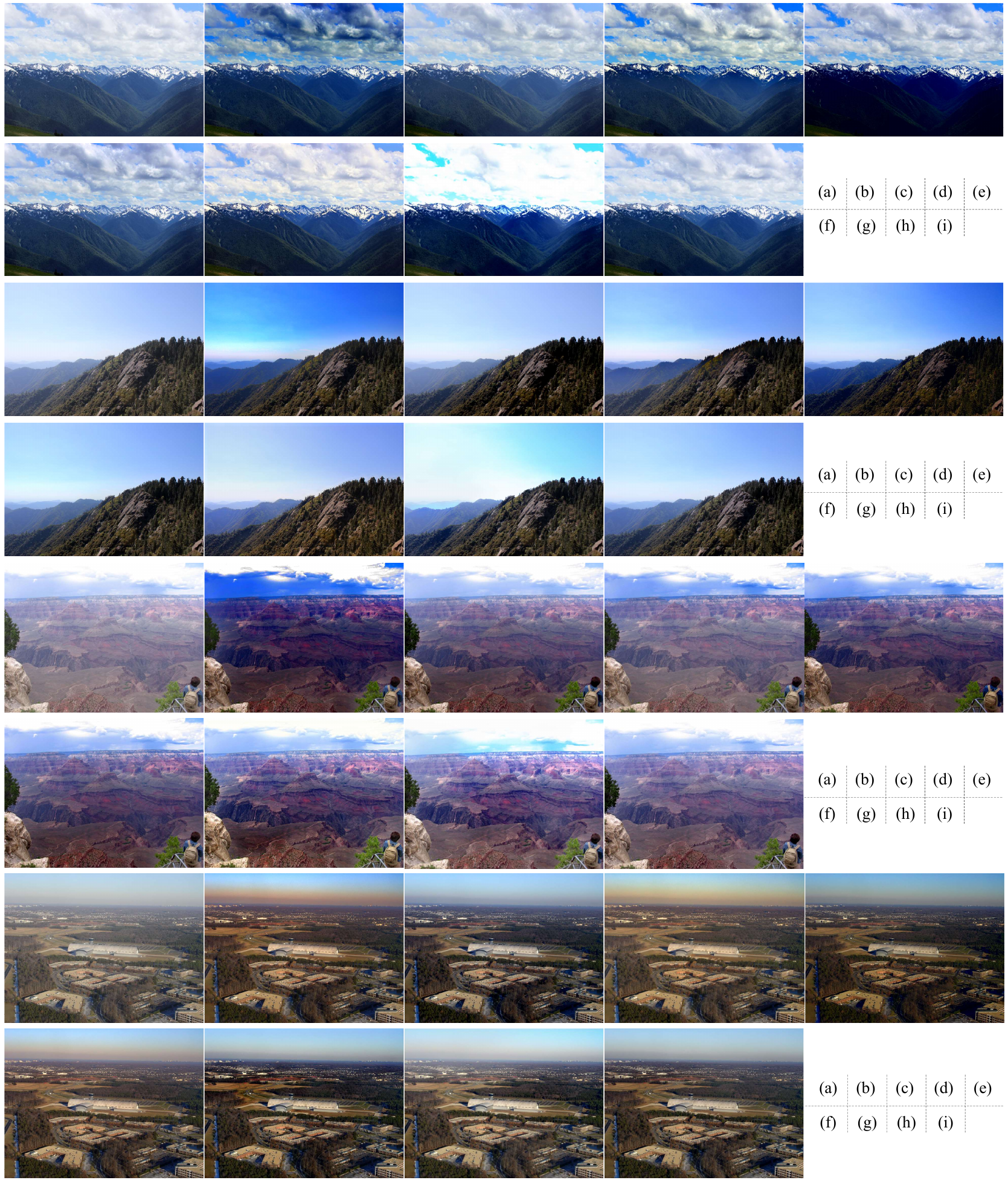}
\caption{Subjective comparisons between FAMED-Net and state-of-the-art methods on real-world hazy images. Best viewed in color. (a) Hazy images. (b) DCP \cite{CVPR2009_He}. (c) DehazeNet \cite{cai2016dehazenet}. (d) MSCNN \cite{ren2016single}. (e) AOD-Net \cite{li2017all}. (f) FPCNet \cite{Zhang2018Fully}. (g) GFN \cite{Ren_2018_CVPR}. (h) DCPDN \cite{zhang2018densely}. (i) FAMED-Net.}
\label{fig:subjective_visual_supp1}
\end{figure*}

\begin{figure*}
\centering
\includegraphics[width=1\linewidth]{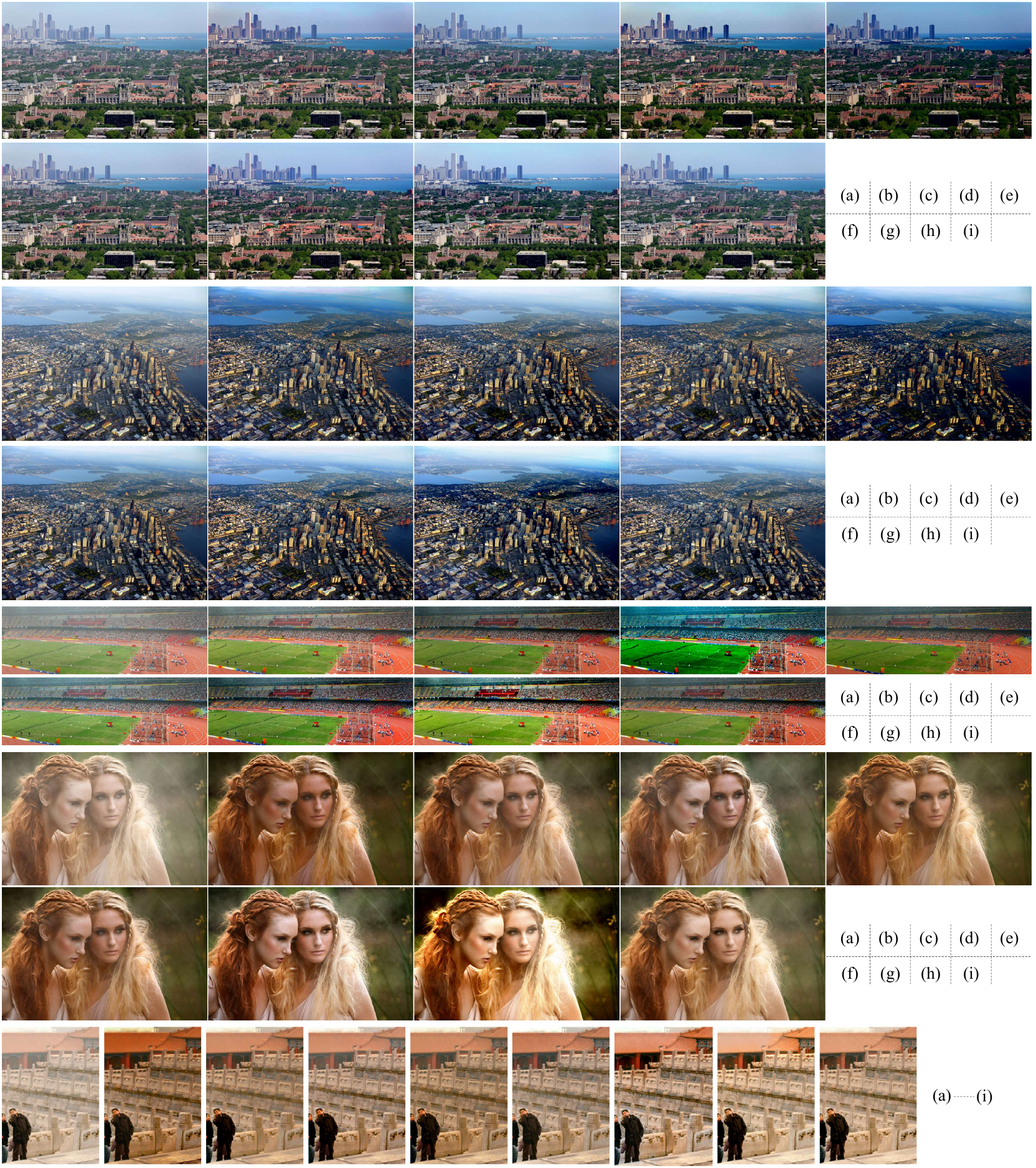}
\caption{Subjective comparisons between FAMED-Net and state-of-the-art methods on real-world hazy images. Best viewed in color. (a) Hazy images. (b) DCP \cite{CVPR2009_He}. (c) DehazeNet \cite{cai2016dehazenet}. (d) MSCNN \cite{ren2016single}. (e) AOD-Net \cite{li2017all}. (f) FPCNet \cite{Zhang2018Fully}. (g) GFN \cite{Ren_2018_CVPR}. (h) DCPDN \cite{zhang2018densely}. (i) FAMED-Net.}
\label{fig:subjective_visual_supp2}
\end{figure*}

More subjective comparisons of FAMED-Net and several state-of-the-art methods including DCP \cite{CVPR2009_He}, DehazeNet \cite{cai2016dehazenet}, MSCNN \cite{ren2016single}, AOD-Net \cite{li2017all}, FPCNet \cite{Zhang2018Fully}, GFN \cite{Ren_2018_CVPR} and DCPDN \cite{zhang2018densely} on real-world hazy images are shown in Figure~\ref{fig:subjective_visual_supp1} and Figure~\ref{fig:subjective_visual_supp2}. As can be seen, FAMED-Net produced better or at least comparable results to state-of-the-art methods with clear details, less color artifacts, and high fidelity in sky regions.

\begin{figure*}
\centering
\includegraphics[width=1\linewidth]{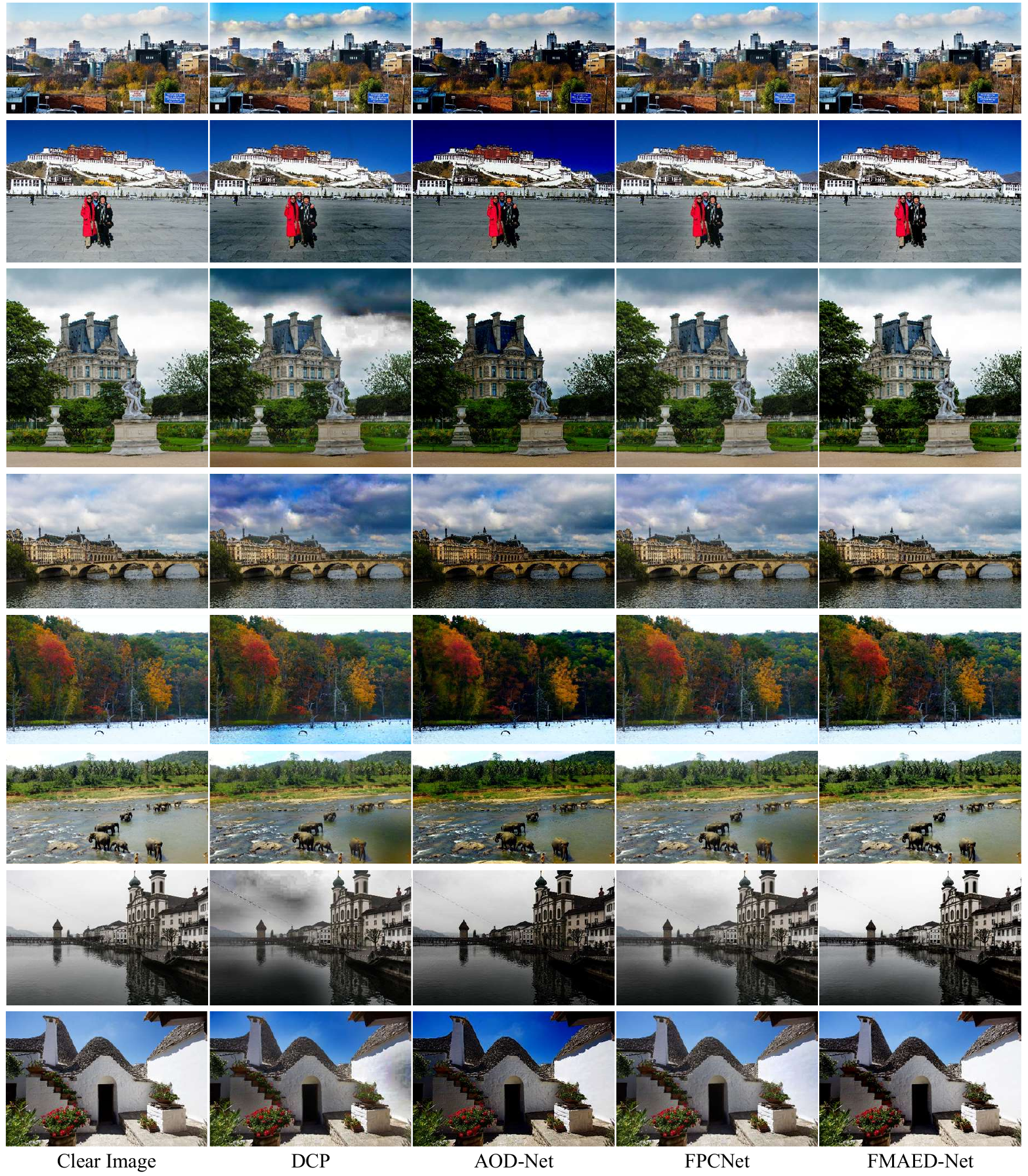}
\caption{Dehazed results of DCP \cite{CVPR2009_He}, AOD-Net \cite{li2017all}, FPCNet \cite{Zhang2018Fully} and FAMED-Net on haze-free images.}
\label{fig:clearDehazed_supp}
\end{figure*}

More subjective comparisons of FAMED-Net and DCP \cite{CVPR2009_He}, AOD-Net \cite{li2017all} and FPCNet \cite{Zhang2018Fully} on haze-free images are shown in Figure~\ref{fig:clearDehazed_supp}. These results demonstrate that FAMED-Net learned a much effective statistical regularity than DCP, FPCNet and AOD-Net. Please refer Section V-C-4 in the paper for more details.\\

\end{document}